\documentclass[10pt]{article} 
\usepackage[preprint]{tmlr}

\usepackage{amsmath,amsfonts,bm}

\def\eqref#1{equation~\ref{#1}}

\def\1{\bm{1}}

\DeclareMathAlphabet{\mathsfit}{\encodingdefault}{\sfdefault}{m}{sl}
\SetMathAlphabet{\mathsfit}{bold}{\encodingdefault}{\sfdefault}{bx}{n}

\usepackage{hyperref}
\usepackage{url}
\usepackage{placeins}
\usepackage{calc}
\usepackage{float}
\usepackage{graphicx}

\title{Scaffold Effects on GAIA: A Controlled Comparison}

\author{\name Jason Starace \email jstarace@mac.com \\
      \addr Independent Researcher
}

\def\month{MM}  
\def\year{YYYY} 
\def\openreview{\url{https://openreview.net/forum?id=XXXX}} 

\begin{document}

\maketitle

\begin{abstract}
    Published agent capability scores conflate what a model can do with what its scaffold lets it do, and the magnitude of this elicitation gap is not well characterized under controlled conditions. This study executes a pre-registered controlled comparison of three scaffolds (ReAct, a Planner-Actor-Rater multi-agent design, and planner-then-executor) across five models from three providers (Claude Opus 4.7, Sonnet 4.6, Haiku 4.5; Gemini 3.1 Pro Preview; GPT-5.5) on GAIA validation Levels 1 and 2, holding tasks and conditions fixed, with three attempts per question. Scaffold choice alone moves measured accuracy by as much as 28 percentage points within a single model (Opus, Level 2, robust slice), confirming the pre-registered hypothesis that scaffold variation produces gaps of at least 10 points. The pre-registered prediction that more capable models would be less scaffold-sensitive is rejected in direction: scaffold effects vary significantly by model in every dataset slice, but the most capable Anthropic model gains the most from structured scaffolds at the harder level, and tier-scaling holds only at Level 1 under the robust slice. The multi-agent advantage over ReAct at Level 2 appears within the Anthropic family but not for the cross-provider models, making model family rather than capability tier the conditioning variable, and the predicted planner-executor advantage on file-reading tasks is falsified. Structured scaffolds make fewer tool calls yet recover more often from mid-trajectory errors at the harder level, and a single cell (Gemini with planner-then-executor) is the cheapest at both levels and the most accurate at Level 2. These results indicate that single-scaffold capability numbers are scaffold-conditional estimates and that the elicitation gap is not guaranteed to shrink as models improve.
\end{abstract}

\section{Introduction}

    Frontier model evaluations treat reported performance as measured capability while acknowledging the measurement is bounded by elicitation quality~\citep{pimpale2025,gringras2026,kapoor2026,barnett2024}; the METR time horizon paper~\citep{kwa2026measuring} discusses this in its limitations appendix, noting that proper elicitation can make a very large difference in performance and treating its published numbers as reasonable lower bounds. As models grow more capable, the magnitude of this gap becomes increasingly consequential for evaluation methodology.
    
    Published capability scores therefore conflate two things: what the model can do, and what the scaffold lets it do. When one team measures Model A with their best scaffold and another measures Model B with theirs, the comparison is confounded by scaffold quality, and capability extrapolations and time-horizon forecasts built on these numbers silently embed the confound.

    To our knowledge, no controlled study has measured the magnitude of scaffold effects on GAIA across a current frontier lineup; adjacent work is discussed in Section~\ref{sec:related_work}. This study addresses that gap, evaluating three scaffolds (Section~\ref{subsec:scaffolds}) across five models, three flagship frontier models (Claude Opus 4.7, GPT-5.5, Gemini 3.1 Pro Preview) and two additional Anthropic-family models (Claude Sonnet 4.6, Claude Haiku 4.5), on GAIA validation Levels 1 and 2. GAIA is chosen because its tasks require an agent loop, making it an appropriate benchmark for scaffold comparison.

    We make three contributions: a controlled comparison isolating scaffold effects (three scaffolds, five models, GAIA Levels 1 and 2, with tasks and conditions held fixed across cells), quantifying the elicitation impact named but not characterized in METR's framework~\citep{kwa2026measuring}; per-cell accuracy estimates with bootstrap confidence intervals and a mixed-effects logistic regression testing the scaffold-by-model interaction, providing statistical claims about scaffold sensitivity rather than point estimates alone; and public release of code, configuration files\footnote{Anonymized repository: \url{https://anonymous.4open.science/r/gaia-scaffold-comparison-ECBF/}}, per-attempt evaluation logs\footnote{Link available on acceptance.}, and the pre-registered sample ID snapshot \citep{starace2026prereg}, enabling full reproduction of the analysis pipeline.

    \subsection{Research questions and hypotheses}

    This study tests the hypotheses and addresses the research questions pre-registered by Starace \citep{starace2026prereg}. The pre-registered experiment split the questions into primary (1-3) and secondary (4-5). Since all five are addressed in this study, they are presented as a flat list.

        \subsubsection{Research questions}
        \begin{enumerate}
            \item \textbf{RQ1:} How much does measured GAIA accuracy change as scaffold varies, holding model and task fixed?
            \item \textbf{RQ2:} Does scaffold sensitivity vary by model? Specifically, does the gap between best-scaffold and worst-scaffold accuracy shrink as model capability grows, or does it persist?
            \item \textbf{RQ3:} What behavioral signatures (action count, tokens emitted, time-to-completion, recovery-from-failure rate) accompany scaffold-driven accuracy differences?
            \item \textbf{RQ4:} How does cost-per-correct-answer compare across scaffold-model combinations?
            \item \textbf{RQ5:} Within the Claude 4.x family (Haiku, Sonnet, Opus), does scaffold sensitivity scale predictably with model tier?
        \end{enumerate}
    
        \subsubsection{Hypotheses}
        \begin{itemize}
            \item \textbf{H1: }Scaffold variation produces at least a 10 percentage-point gap between best and worst scaffold for at least one model on GAIA Level 1.
            \item \textbf{H2: }Frontier models (Opus 4.7, Gemini 3.1 Pro Preview, GPT-5.5) will show smaller scaffold gaps than mid-tier models, consistent with the hypothesis that elicitation matters less when the model is more capable.
            \item \textbf{H3: }Triframe-style multi-agent scaffolds will outperform single-agent ReAct on GAIA Level 2 but not on Level 1.
            \item \textbf{H4: }Planner-then-executor scaffolds will outperform on tasks with file-reading components, where planning a sequence of operations matters.
        \end{itemize}   

\section{Related work}\label{sec:related_work}

    Scaffold variation has appeared frequently in agent evaluation work, but typically as an instrument rather than the object of study. \citet{yao2023react} introduce ReAct, the alternating reason-act loop that is now a standard agent scaffold; AgentBench~\citep{liu2025agentbench} evaluates LLMs as agents across eight environments while holding the agent loop roughly fixed; \citet{snell2024scaling} examine inference-time compute allocation rather than loop structure. Across this line, scaffold is part of the apparatus, not the manipulated variable.

    METR's time horizon framework~\citep{kwa2026measuring} is the canonical statement of the elicitation-as-lower-bound problem motivating this study: its limitations appendix (E.4) treats measured horizons as elicitation-bounded lower bounds, and the paper spent two to three engineer-weeks tuning scaffolds for two of its models with lighter elicitation for the rest of its sweep. It names the elicitation gap and acknowledges its size matters for interpretation, but does not characterize the gap through controlled comparison on a shared benchmark.

    Recent work quantifies scaffold and elicitation effects in adjacent settings. \citet{pimpale2025} forecast frontier agent capabilities under a low-versus-high elicitation split, reporting 33\% versus 62.2\% for Claude Sonnet 3.5 on SWE-Bench Verified, a roughly 30 percentage-point gap on a single model-benchmark pair. \citet{ndzomga2026} evaluate 33 scaffolds and 70+ model configurations across eight benchmarks, finding absolute score prediction degrades under scaffold-driven shift while rank order remains stable. \citet{gringras2026safety} report one of the largest controlled scaffold studies on safety benchmarks (N = 62{,}808; six frontier models, four deployment configurations), finding scaffold effects on measured safety are small once evaluation format is held constant. At the publication level, \citet{gringras2026} audit 4{,}766 LLM evaluation papers and find a median publication elicitation gap of 10.85 ECI points behind the frontier at evaluation time: where the present study measures how much scaffold choice shifts a single model's measured capability, \citet{gringras2026} measure how much published claims drift from frontier conditions through reporting practices.

    This study is a controlled replication. The scaffold-matters claim is not novel: by 2026 the conventional wisdom in agent evaluation is that scaffold choice produces meaningful performance differences. What remains under-specified is the magnitude under controlled comparison on a general-assistant benchmark, whether it scales predictably with model capability, and how it interacts with task difficulty. Prior controlled scaffold work covers rank preservation, safety, and elicitation forecasts on coding benchmarks; none addresses general-assistant capability evaluation across a current frontier lineup. This study does: three scaffolds, five models including the full Anthropic Opus/Sonnet/Haiku ladder, GAIA Levels 1 and 2, with hypotheses and design pre-registered by Starace prior to data collection~\citep{starace2026prereg}.

\section{Methodology}

    \subsection{Testbed}\label{subsec:testbed}
        The testbed is GAIA~\citep{mialon2023gaia}, a benchmark of 466 general-assistant tasks requiring multi-modality handling, web browsing, file reading, and multi-step reasoning. Tasks are stratified into three levels of difficulty. Validation answers are public; test answers are held by the Hugging Face leaderboard. This study evaluates validation Levels 1 and 2; Level 3 is excluded for scope.
    
    \subsection{Evaluation framework}\label{subsec:framework}
    
        Evaluations are implemented using Inspect AI~\citep{aisi2024inspect}, an open-source framework for large language model evaluations developed by the UK AI Security Institute. The companion package \texttt{inspect\_evals}~\citep{aisi2024inspectevals} provides reference implementations of common benchmarks, including built-in support for GAIA at task granularity (\texttt{gaia\_level1}, \texttt{gaia\_level2}, \texttt{gaia\_level3}) and a working ReAct baseline. The official GAIA scorer is included in the implementation and used directly without modification; no model-graded scoring is used.
        
        Inspect 0.3.217 did not yet pass \texttt{ToolConfig.include\_server\_side\_tool\_invocations=True} when combining Gemini's native \texttt{google\_search} with function tools, so a local patch was applied to the provider for the duration of this study. The patch enables the same tool combination across all five models, preserving like-for-like comparison. Full patch details are in Appendix~\ref{app:gemini_patch}.
        
    \subsection{Scaffolds}\label{subsec:scaffolds}

        Three scaffolds are evaluated, all implemented in Inspect AI. S2 and S3 share an identical tool set: web search (via Inspect's \texttt{web\_search} tool, which routes to provider-native search where available), a sandboxed shell (\texttt{bash}), a Python sandbox (\texttt{python}), and a text editor (\texttt{text\_editor}). S1, the unmodified \texttt{inspect\_evals} baseline, carries that implementation's surface instead: \texttt{bash}, \texttt{python}, and Inspect's stateful \texttt{web\_browser} (Section~\ref{sec:limitations}).
        
        \textit{S1: ReAct (baseline).} The standard single-agent loop. The model receives the task and plans and acts in alternation, with tool access for browsing, code execution, and file reading. This is the \texttt{inspect\_evals} default GAIA solver and acts as the baseline against which the other scaffolds are compared. The default ReAct prompt from \texttt{inspect\_evals} is used without modification.
        
        \textit{S2: Planner-Actor-Rater (PAR).} Three roles share a task state: a planner that decomposes the task, an actor that executes individual subtasks with tools, and a rater that reviews progress and signals completion or revision. The pre-registration named this scaffold ``triframe-style multi-agent''; the rename and its rationale are documented in Section~\ref{subsec:deviations}. Full role prompts are provided in Appendix~\ref{app:par_prompts}.
        
        \textit{S3: Planner-then-Executor.} Two phases: an explicit planning step where the model produces a written plan with no tool access, followed by an execution phase where a fresh prompt receives the plan and executes against tools. S3 tests whether separating planning from execution helps relative to interleaved planning and action. The same model is used for both phases; cross-model variants are out of scope, as varying the model within a scaffold would confound scaffold effect with model-mixing effect. Full phase prompts are provided in Appendix~\ref{app:pte_prompts}.
    \subsection{Models}\label{subsec:models}
    
        Five models from three providers are evaluated: Claude Opus 4.7, Claude Sonnet 4.6, and Claude Haiku 4.5 (Anthropic); Gemini 3.1 Pro Preview (Google); and GPT-5.5 (OpenAI). The Anthropic ladder (Opus, Sonnet, Haiku) is the primary analysis vehicle for RQ5 (whether scaffold sensitivity scales predictably with model tier within a family). The cross-provider entries (Gemini, GPT-5.5) provide flagship-to-flagship checks for the main findings.
    
        Model IDs are pinned to dated snapshots where the provider treats the dateless ID as an alias (Haiku 4.5, GPT-5.5). Anthropic 4.6+ models use the dateless canonical ID, which is itself a pinned snapshot. Gemini preview models cannot be pinned beyond the preview string and carry a reproducibility risk that is documented in the limitations.
    
        Reasoning mode is not enabled for any model; default response generation is used across all providers. This deviates from the pre-registration and is documented in Section~\ref{subsec:deviations}; the controlled-comparison structure is unaffected because reasoning is held constant (off) across all cells.

    \subsection{Experimental design}\label{subsec:experimental_design}
    
        \textit{Design and attempts.} The experimental design is a 3 (scaffold) $\times$ 5 (model) factorial across GAIA validation Levels 1 and 2. Each cell is run with three attempts per question to support per-cell variance estimation. With 53 questions at Level 1 and 86 at Level 2, the full design specifies $15 \times 139 \times 3 = 6{,}255$ validation runs.
        
        \textit{Metrics.} For each (model, scaffold, question, attempt), the following are recorded: binary accuracy from GAIA's official scorer, the primary outcome for RQ1 and RQ2; action count (tool calls executed), total tokens, and wall-clock time to completion, supporting RQ3's behavioral signatures; and cost in dollars per attempt, computed from token counts at provider rates (hereafter \emph{log-derived cost}, distinct from the billed amount, a gap discussed in Section~\ref{sec:limitations}), feeding RQ4's cost-per-correct ranking.
        
        \textit{Statistical analysis.} The confirmatory analyses follow the pre-registered plan: per-cell bootstrap accuracy with 95\% CIs, the max-minus-min scaffold gap per (model, level) as the primary statistic for H1 and H2, a mixed-effects logistic regression with a scaffold $\times$ model interaction as the inferential test for H2, and dollars per correct answer ranked across cells for RQ4. Each analysis is specified in full in its results subsection.
        
        \textit{Reproducibility infrastructure.} A wrapper script re-runs only samples without a clean completion across prior \texttt{.eval} logs, retrying Inspect-level errors and unexecuted samples but never wrong answers, and merges multiple logs per shard at analysis time. The ordered list of GAIA validation task IDs per level, generated from a direct \texttt{load\_dataset()} call against \texttt{gaia-benchmark/GAIA} (\texttt{2023\_all}, validation split), is committed to \texttt{configs/gaia\_sample\_ids.json}, making the dataset-to-position mapping part of the experimental record.
        
    \subsection{Deviations from pre-registration}\label{subsec:deviations}

        This subsection documents deviations from the pre-registered methodology~\citep{starace2026prereg}; Section 3.8 of the pre-registration licenses prompt modifications for unambiguous implementation defects.
        
        \textit{Reasoning mode.} The pre-registration specified reasoning mode enabled, with default budget, for all models supporting it. The runs applied no reasoning configuration, which Inspect AI treats as disabled: pilot work showed reasoning costs scaling unpredictably across providers, and reasoning off preserved the comparison under budget constraints. The comparison structure is unaffected, since reasoning is constant (off) across all cells.
        
        \textit{Scaffold S2 renamed.} The pre-registration named S2 ``triframe-style multi-agent,'' modeled on METR's triframe agent~\citep{kwa2026measuring}, while noting the implementation is an approximation rather than a faithful reproduction. The three-role design (planner, actor, rater) differs substantively from METR's triframe, a single-model plan-and-vote loop, and is renamed Planner-Actor-Rater (PAR); the pre-registered H3 label ``triframe-style'' refers to the same scaffold.

        \textit{Input truncation.} Accumulated scaffold context occasionally exceeded model context windows, most notably the 200K ceilings of Claude Haiku and Sonnet. Input truncation at 200K tokens, dropping the oldest tool-call traces first, was implemented under the Section 3.8 license. It applies to the rerun of failed cells only; cells completed before it was added are unaffected.

        \textit{Tool surface.} As run, S1 retained the unmodified \texttt{inspect\_evals} tool set rather than the constant surface specified in the pre-registration; the resulting confound in s1 contrasts is discussed in Section~\ref{sec:limitations}.
        
        \textit{Incomplete cells.} The full design specifies 6{,}255 validation runs (3 attempts $\times$ 5 models $\times$ 3 scaffolds $\times$ 139 questions). The validation run completed 5{,}907 of these; 15 of the 30 (model, scaffold, level) cells are partial, driven primarily by the provider-side serialization defect described in Appendix~\ref{sec:provider_bug}, with smaller contributions from context-overflow, content-filter, and PDF page-limit failures. Affected cells are listed in Appendix~\ref{app:incomplete_cells}. Analyses run on completed attempts only. The pre-registered held-out test-set submission was not executed; Section~\ref{sec:limitation_contamination} discusses the resulting contamination control.

\section{Results}
    Sections~\ref{sec:scaffold_gap} and~\ref{sec:pairwise_hypotheses} present the pre-registered confirmatory analyses testing H1 through H4.. Each analysis is reported under three slices of the dataset. The \emph{primary} slice includes every attempt-unit and treats errored units as incorrect, matching the headline accuracy denominator used elsewhere. The \emph{robust (no flag)} slice excludes units carrying the \texttt{provider\_serialization\_bug} flag described in Appendix~\ref{sec:provider_bug}, isolating scaffold-attributable variation from the SDK-level defect. The \emph{intersection} slice retains only sample\_ids for which every (model, scaffold) cell at that level had at least one attempt-unit complete without an Inspect-level error, controlling for question-set asymmetry on a shared question pool.

    \subsection{Scaffold gap (H1 and H2)}\label{sec:scaffold_gap}

        \subsubsection{Per-cell outcomes}\label{sec:per_cell_outcomes}
            
            Each (model, scaffold, level) cell was attempted three times per sample\_id under the design described in Section~\ref{subsec:experimental_design}, giving 159 attempt-units per L1 cell and 258 per L2 cell. Per-cell accuracy is visualized in Figure~\ref{fig:per_cell_accuracy} (Appendix~\ref{app:bootstrap_accuracy_full}); completion rates, accuracy, and failure-mode bucket counts for all five models are reported per cell in Table~\ref{tab:per_cell_outcomes_full} (Appendix~\ref{app:per_cell_outcomes_full}). Two accuracy columns are reported there: \emph{accuracy overall}, dividing correct completions by the full attempt count with errored units counted incorrect, and \emph{accuracy on completed}, dividing by non-errored units only; the columns diverge only where provider-side issues prevented completion of some attempts. Completion rates are at 1.00 for every cross-provider cell. Within the Anthropic ladder, the largest gaps belong to Opus (0.82 s2 L2, 0.79 s3 L2, 0.94 s2 L1, 0.84 s3 L1), driven almost entirely by \texttt{provider\_serialization\_bug}-flagged units (Appendix~\ref{app:provider_bug_summary}), and to Haiku on s2 and s3 L2 (0.81 and 0.88), driven primarily by the same flag with a secondary \texttt{prompt\_too\_long} contribution (Appendix~\ref{sec:prompt_too_long}). Accuracy on completed is in most cases comparable to or higher than accuracy overall, indicating that the errored units would not, on average, have raised the cell's headline number had they completed cleanly.

        \subsubsection{Bootstrap accuracy}\label{sec:bootstrap_accuracy}
            
            Confidence intervals around each per-cell point estimate are computed by clustered bootstrap: sample\_ids are resampled with replacement and an included sample\_id contributes all of its attempt-units as a block, preserving the within-question dependence of the three-attempts design. 5{,}000 resamples per cell with a fixed seed give the 95\% interval from the 2.5th and 97.5th percentiles; the seed is shared across the three slices so differences between slices reflect only the input pool. The per-cell dot plot (Figure~\ref{fig:per_cell_accuracy}, Appendix~\ref{app:bootstrap_accuracy_full}) makes two patterns visible. The cross-provider CIs overlap heavily across scaffolds, with one exception: Gemini L2, where the s3 interval sits clear of s1. The within-Anthropic ladder shows non-overlapping CIs for Haiku L1 (s3 separates from s1 and s2) and for Sonnet and Opus L2 under the robust slice (s1 separates from s2 and s3).

        \subsubsection{Scaffold gap statistic}\label{sec:scaffold_gap_stat}
        
\begin{figure}
                \centering
                \includegraphics[width=.75\textwidth]{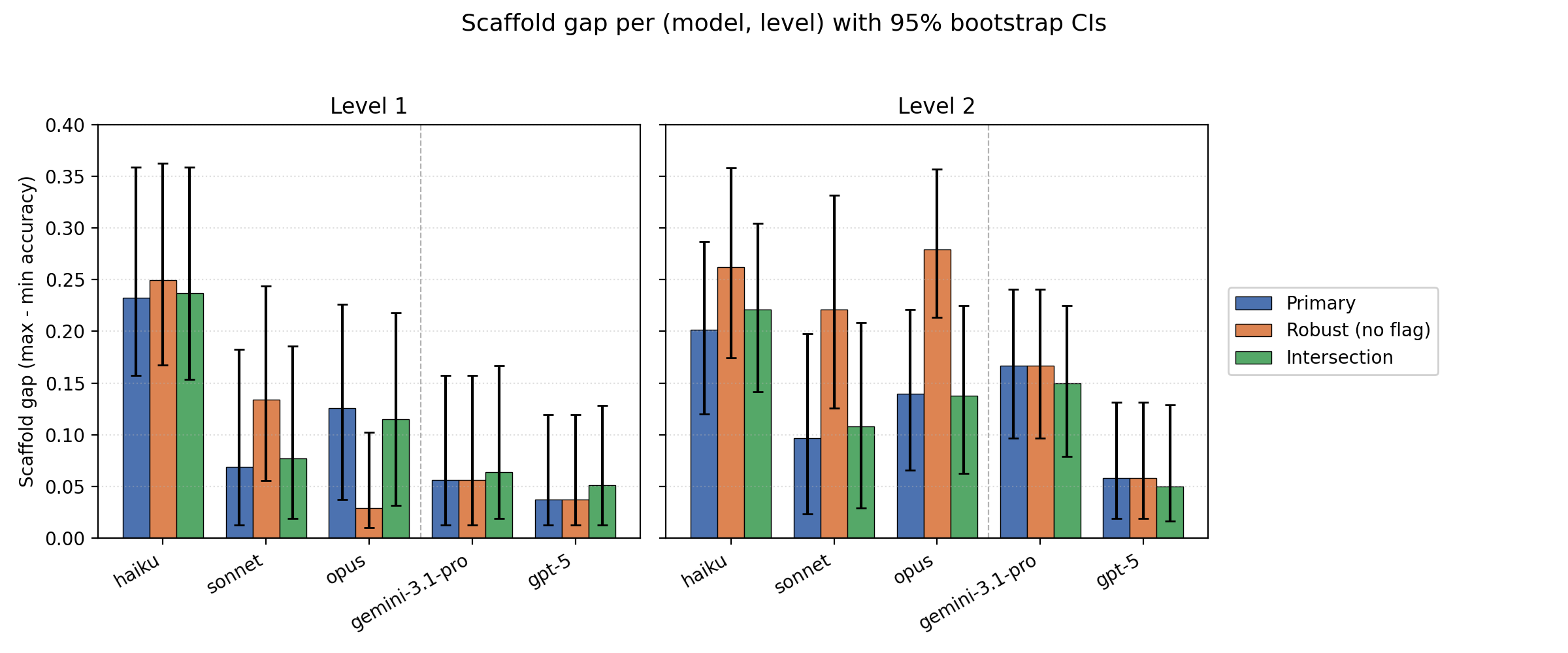}
                \caption{Scaffold gap (max minus min accuracy across the three scaffolds) per (model, level) with 95\% bootstrap CIs, shown for the primary, robust, and intersection slices. The dashed line separates the Anthropic ladder from the cross-provider models.}
                \label{fig:scaffold_gap_bars}
            \end{figure}

            The scaffold gap is the maximum minus the minimum accuracy across the three scaffolds for a given (model, level) cell. Its CI extends the per-cell bootstrap: in each of the 5{,}000 iterations, sample\_ids are resampled once per (model, level) and the same picks feed all three scaffold accuracies, so the within-question dependence across scaffolds is preserved and the gap CI is exact rather than a propagated independent-CI approximation. Figure~\ref{fig:scaffold_gap_bars} displays the gap and its CI for each (model, level, slice) combination. The full headline gap tables across all three slices, including which scaffold sits at the max and min, are in Appendix~\ref{app:scaffold_gap_full}; the complete pairwise contrasts (s2$-$s1, s3$-$s1, s3$-$s2) are in Appendix~\ref{app:scaffold_gap_pairwise}.

            At Level 1, the gap is small for the three frontier models (Opus 0.13, Gemini 0.06, GPT-5 0.04, primary slice) and largest for Haiku (0.23). At Level 2, the primary-slice gaps are 0.20 for Haiku, 0.10 for Sonnet, 0.14 for Opus, 0.17 for Gemini, and 0.06 for GPT-5. Under the primary slice the Anthropic ladder is non-monotonic in tier at both levels: Sonnet, not Opus, is the least scaffold-sensitive (Appendix~\ref{app:tier_sensitivity}).
    
            The robust slice changes this picture for the Anthropic models at Level 2. Removing flagged units expands the L2 gap for Sonnet (0.10 $\to$ 0.22) and Opus (0.14 $\to$ 0.28) while leaving the cross-provider gaps unchanged: the SDK defect selectively suppressed s2 and s3 accuracies, so removing it widens rather than shrinks the visible scaffold gaps. Opus L1 inverts in the opposite direction (0.13 $\to$ 0.03) because s1 was Opus's strongest L1 scaffold, so the same mechanism collapses that gap instead. The intersection slice tracks the primary slice closely at both levels, consistent with the small reduction in retained sample\_ids.
    
            Read against H1, the L2 results give clear positive evidence under all three slices for Haiku, Sonnet, and Opus, weaker positive evidence for Gemini, and a borderline case for GPT-5, whose gap CIs exclude zero by a narrow margin. H1 is supported but the magnitude is highly model-dependent. The within-Anthropic ladder is taken up in Section~\ref{sec:tier_sensitivity}.

        \subsubsection{Mixed-effects test of H2}\label{sec:h2_inferential}
        
            H2 predicted that less-capable models would show larger scaffold effects. This analysis adds the inferential test pre-registered in Section~3.6 of the pre-registration: a logistic mixed-effects model with success as the binary outcome, sample\_id as a random intercept, and scaffold, model, and level as fixed effects, fit via lme4's \texttt{glmer} under the pymer4 wrapper with reference levels s1 and Opus. The H2 test is the likelihood-ratio test comparing a no-interaction baseline against a full scaffold $\times$ model interaction. The interaction model's marginally elevated gradient at convergence (0.001 to 0.013 across slices) is above lme4's strict threshold but well within practically converged range; we treat the fits as usable for inference. Full coefficient tables and fit metadata are in Appendix~\ref{app:glmm_full}.
        
            The LRT rejects the null of no interaction in every slice ($\chi^2(8) = 82.95$ primary, $122.18$ robust, $81.38$ intersection; all $p < 0.0001$): scaffold effects are statistically distinguishable across models. This part of H2 holds.
            
            The directionality of the interaction does not match H2's prediction that scaffold effects shrink as capability increases. Under the robust slice the s2 and s3 main effects for the Opus reference are strongly positive (+1.20 and +1.29 log-odds), while most scaffold:model interactions for the non-Opus models are negative and significant (full coefficients in Appendix~\ref{app:glmm_full}): read against the Opus reference, the other models gain \emph{less} from s2 and s3 than Opus does. H2's directional claim is falsified. The descriptive counterpart makes the direction concrete: the largest robust-slice gap in the experiment belongs to Opus L2 (0.28), the most capable Anthropic model, while Haiku's comparable gap (0.26) reflects a baseline floor that s2 and s3 do not lift it out of, which the model captures as Haiku's negative interactions.
            
            Read against H2 specifically: the inferential test rejects the null, but the direction of variation does not support the tier-scaling claim. The data are consistent with a more capability-permissive scaffold benefit: Opus gains the most from structured scaffolds at the harder level. The conservative reading is that H2's prediction holds at L1 under the robust slice and fails at L2 under the same slice; the pooled mixed-effects model reports the L2-dominated falsification because L2 has more observations and larger gap magnitudes.

    \subsection{Pairwise hypothesis tests (H3 and H4)}\label{sec:pairwise_hypotheses}

        \subsubsection{Triframe versus ReAct (H3)}
            
            H3 predicted that triframe-style multi-agent scaffolds (s2) would outperform single-agent ReAct (s1) on GAIA Level 2 but not on Level 1. The test statistic is the per-(model, level, slice) accuracy difference acc\_s2 minus acc\_s1 with 95\% sample\_id-clustered bootstrap CI; the full contrasts are tabulated in Appendix~\ref{app:scaffold_gap_pairwise}, where the pair is reported as s1$-$s2, the negation of the H3 contrast. A positive s2 minus s1 value means triframe outperformed ReAct; the H3 prediction is that L2 cells are positive and L1 cells are not.
        
            Under the primary slice, the L1 prediction holds in the weak sense that no L1 cell shows a triframe advantage with a CI strictly above zero; Opus L1 is the most negative (-0.069) and Sonnet L1 the most positive (+0.069). The L2 prediction holds in part: all five models show positive point estimates, but only Haiku (+0.112, CI [0.023, 0.202]) and Opus (+0.140, CI [0.062, 0.221]) have CIs strictly above zero, while Gemini (+0.058) and GPT-5 (+0.016) span zero. Triframe provides a meaningful L2 advantage for the Anthropic family but not for the cross-provider models.
        
            The robust slice makes the L2 finding considerably stronger inside the Anthropic family: removing the \texttt{provider\_serialization\_bug} units, the s2 minus s1 estimates at L2 are +0.193 (Haiku), +0.221 (Sonnet), and +0.279 (Opus), all with CIs strictly above zero, while the cross-provider L2 estimates are unchanged because the flag did not affect them. At L1 under the robust slice, Sonnet is the lone CI strictly above zero (+0.129 [0.030, 0.232]); the other four models are consistent with the L1 prediction.
            
            Read against H3 specifically: the prediction is supported within the Anthropic family at L2, holds at L1 except the robust-slice Sonnet cell, and is not supported for the cross-provider models at L2. H3 is therefore conditionally supported, with the conditioning variable being model family rather than capability tier per se.

        \subsubsection{Planner-executor on file-reading tasks (H4)}\label{sec:h4}
        
            H4 predicted that planner-then-executor scaffolds (s3) would outperform on tasks with file-reading components. We operationalize file-reading tasks as the document-attachment subset of GAIA validation: tasks shipping with .xlsx, .pdf, .docx, .pptx, .txt, or .csv attachments per the dataset metadata (six L1 and ten L2 sample\_ids). The H4 test is the difference-of-differences on s3 minus s1: the s3-over-s1 advantage on the file subset minus the s3-over-s1 advantage on the non-file complement. A positive DoD with CI strictly above zero supports H4 at that cell; a negative DoD with CI strictly below zero contradicts it.

            Across the ten (model, level) cells under the primary slice, the prediction is supported in one (Opus L1, DoD = +0.142, CI [0.030, 0.252]), contradicted in four (Haiku L1, Haiku L2, Gemini L2, GPT-5 L2), and inconclusive in five; the full tables for all slices are in Appendix~\ref{app:h4_robustness}. Two interpretive caveats bound the negative cells. Haiku sits at floor on documents (16.7\% accuracy on L1 documents under s3, 30.0\% under s3 at L2, the lowest cell-and-subset accuracies in the experiment), so its negative DoDs reflect document handling generally rather than a planner-executor-specific failure mode. Gemini and GPT-5 at L2 hit ceiling on the file subset under s1 (both at 100\%), so planning overhead cannot pay off on tasks the model already solves reactively.
            
            None of the three robustness checks rescues H4. Under the robust slice the lone supporting cell disappears: Opus L1 drops to -0.002 (CI [-0.083, 0.086]), attributing the primary-slice positive to flagged units. The intersection slice restores the Opus L1 cell but introduces no others, and the any-file expansion tracks the document-only result closely.
            
            Read against H4 specifically: the prediction is falsified, with one supporting cell out of ten that does not survive the robust check; the floor and ceiling caveats above make the contradicting cells equally unreliable as evidence in the opposite direction.
            
    \subsection{Behavioral signatures (RQ3)}\label{sec:rq3}

        RQ3 asks what behavioral signatures accompany scaffold-driven accuracy differences. We report four per-cell summary statistics under the primary slice: mean action count (tool calls per attempt), mean output tokens emitted, mean working time on completed attempts, and recovery-from-failure rate (fraction of attempts with at least one mid-trajectory tool error that nonetheless produced a correct submission). Output tokens are the headline token statistic: provider-reported totals fold in prompt-caching streams billed and surfaced differently across providers, leaving output the only cross-provider-comparable measure. Full per-cell tables with the input, output, cache-read, cache-write, and total decomposition under all three slices are in Appendix~\ref{app:rq3_behavioral_full}; the input/output split alone is in Appendix~\ref{app:rq3_tokens_split}. CIs are 95\% sample\_id-clustered bootstrap, 5{,}000 resamples, seed shared with the rest of the validation analyses.
        
        Three of the four signatures order the scaffolds monotonically. Averaged across the five models at L2, mean action count is 22.0 under s1, 8.8 under s2, and 6.4 under s3; mean output tokens are 4.8k, 3.1k, and 2.9k; mean working time on completed attempts is 151s, 115s, and 93s. L1 shows the same ordering with compressed values. The reactive baseline dispatches roughly three times the tool calls of planner-then-executor and emits roughly 65\% more output tokens, and the action-count ordering is stable within cells: every model uses fewer actions under s3 than under s1; output tokens follow for every model except Opus, which emits slightly more under s3 at both levels. The per-cell extremes are all reactive-baseline configurations on the harder level (Appendix~\ref{app:rq3_behavioral_full}).
        
        Recovery from failure inverts the pattern. Averaged across models at L2, recovery rate is 0.38 under s1, 0.57 under s2, and 0.51 under s3. At L1 the rates are lower and the inversion does not hold, but four of fifteen L1 cells have no mid-trajectory errors at all (undefined recovery rate), so the L1 averages are noisy. The L2 finding is the substantive one: structured scaffolds make fewer tool calls overall but the in-trajectory errors that do occur are more often recovered from. We do not have a mechanistic claim about why; candidate explanations are taken up in the Discussion. The recovery numbers are descriptive context, not a hypothesis test.
        
        ReAct is thus the most expensive scaffold by every behavioral metric, runs longest, and recovers least often from in-trajectory failure, yet s1 remains competitive on L1 for some models (Opus s1 is the most accurate Opus L1 cell at 0.83): the structural overhead of s2 and s3 frequently does not buy a matching accuracy gain (Section~\ref{sec:scaffold_gap_stat}).
        
    \subsection{Cost per correct}
    \label{sec:cost_per_correct}

        This subsection presents the pre-registered RQ4 analysis: cost-per-correct rankings across the 30 (model, scaffold, level) cells. The headline metric is \emph{realized cost per correct}: the sum of log-derived dollar costs across every attempt-unit in a cell divided by the cell's count of correct completions, with errored units counted as incorrect, matching the headline accuracy denominator (Table~\ref{tab:per_cell_outcomes_full}, Appendix~\ref{app:per_cell_outcomes_full}). A secondary metric, \emph{happy-path cost per correct}, restricts both numerator and denominator to completed units; the two coincide at 100\% completion and diverge where errored units carried token cost. Sample\_id-clustered bootstrap CIs (5{,}000 resamples, seed shared with the Section~\ref{sec:scaffold_gap} analyses) accompany every estimate. Full per-level rankings under all three slices are in Appendix~\ref{app:cost_per_correct_full}; the realized-versus-happy-path comparison is in Appendix~\ref{app:cost_per_correct_happy_path}.
        
\begin{figure}[t]
            \centering
            \includegraphics[width=.75\linewidth]{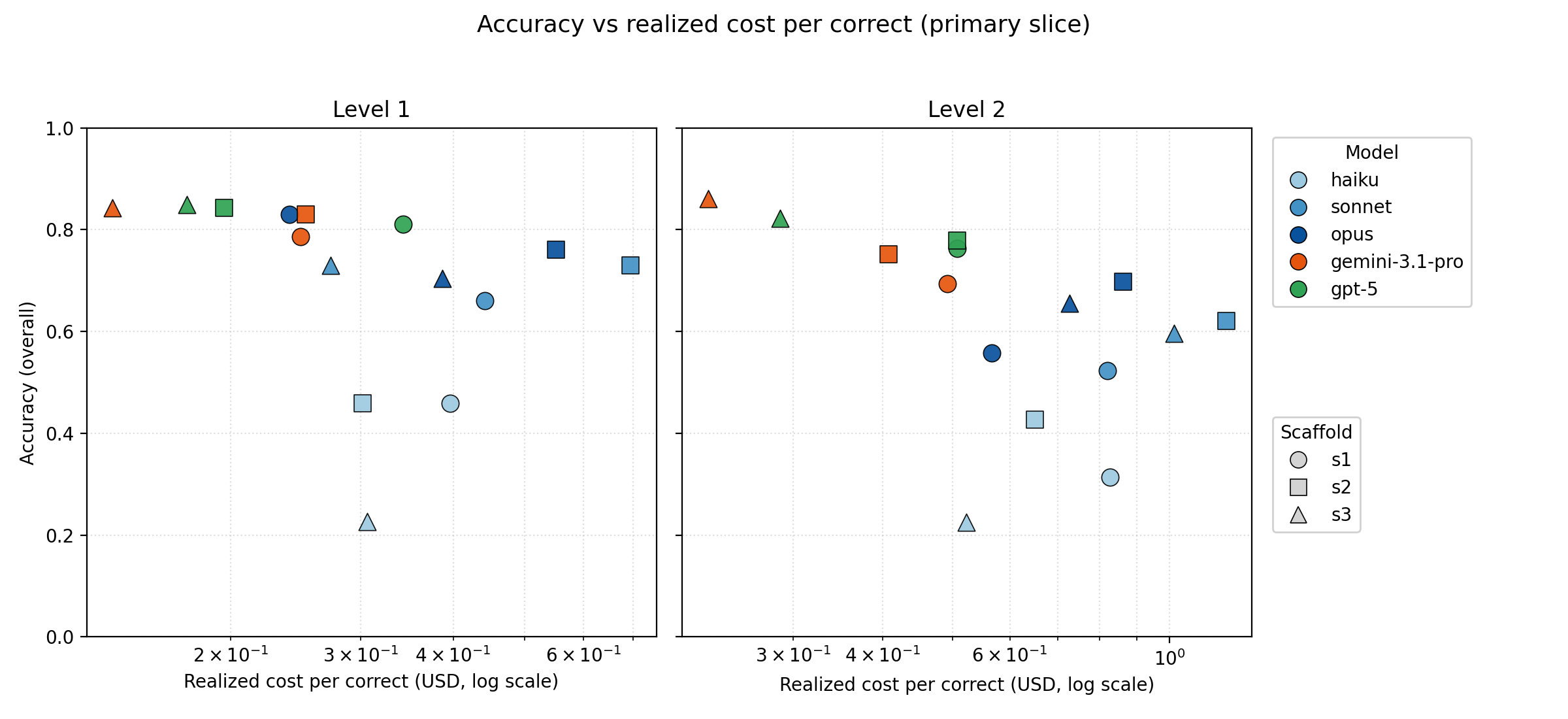}
            \caption{Accuracy versus realized cost per correct for every (model, scaffold, level) cell, primary slice; x-axis logarithmic. Color encodes model; marker shape encodes scaffold.}
            \label{fig:cost_pareto}
        \end{figure}

        Figure~\ref{fig:cost_pareto} plots accuracy against realized cost per correct under the primary slice for every cell. Three cells anchor the L1 Pareto frontier: Gemini s3 (\$0.14/correct at 84.3\% accuracy), GPT-5 s3 (\$0.17 at 84.9\%), and GPT-5 s2 (\$0.20 at 84.3\%). Three anchor the L2 frontier: Gemini s3 (\$0.23 at 86.0\%), GPT-5 s3 (\$0.29 at 82.2\%), and Gemini s2 (\$0.41 at 75.2\%). Gemini s3 is the cheapest cell on both panels and the most accurate at L2; at L1 it concedes 0.6 points to GPT-5 s3 at lower cost, leaving both on the frontier.
        
        Two within-provider L1 features deserve attention. Sonnet s1 is more expensive per correct (\$0.44) than Opus s1 (\$0.24) despite cheaper per-token rates, the product of higher token usage and lower accuracy on the baseline scaffold. And Haiku s3 sits at the bottom of the panel: the accuracy collapse on the planner-executor scaffold (22.6\%) makes Haiku a poor cost-per-correct choice at L1 despite the cheapest per-token rates in the study.
        
        At L2 the spread widens substantially: the cheapest cell (Gemini s3, \$0.23) and the most expensive (Sonnet s2, \$1.20) differ by a factor of 5.2, and the five most expensive L2 cells under the primary slice are all Anthropic. Three of those five are partially driven by units that paid token cost before erroring under the \texttt{provider\_serialization\_bug} pattern (Appendix~\ref{sec:provider_bug}): spend incurred without contributing to the correct denominator. The robust slice (Appendix~\ref{app:cost_per_correct_full}) confirms the inflation, dropping Sonnet s2 L2 from \$1.20 to \$0.97, without changing the qualitative ranking: Sonnet s2 L2 remains the worst cell, the Anthropic L2 cluster remains more expensive than the cross-provider cluster, and Gemini s3 remains the headline winner.
        
        The realized-versus-happy-path comparison (Appendix~\ref{app:cost_per_correct_happy_path}) tracks the same pattern, with the largest divergence at Sonnet s2 L2. Realized remains the conservative procurement estimate: what an operator running the experiment as configured would actually pay per correct answer.
        
        Read against RQ4, the picture is dominated by two findings. First, Gemini s3 is the cheapest cell at both levels and the most accurate at L2 under all three slices, the strongest single cell at the current pricing snapshot. Second, the realized cost-per-correct gap inside the Anthropic family is large at L2: Haiku, Sonnet, and Opus cells span \$0.52 to \$1.20 under the primary slice against \$0.23 to \$0.51 for the comparable cross-provider cells, and the within-Anthropic dispersion is wider than the within-Gemini or within-GPT-5 dispersion, indicating that scaffold choice imposes a larger cost penalty on the Anthropic family at the harder level. The robust slice moves the two findings in opposite directions: removing flagged units expands Anthropic L2 scaffold-accuracy gaps (Section~\ref{sec:scaffold_gap_stat}) while compressing the cost-per-correct spread, since flagged units carried spend without correct answers.

    \subsection{Within-Anthropic tier sensitivity}\label{sec:tier_sensitivity}

        This subsection addresses RQ5: within the Claude 4.x family, does scaffold sensitivity scale predictably with model tier? The robust slice gives a split verdict: the L1 gap ladder is cleanly monotonic and decreasing (0.25, 0.13, 0.03), consistent with the H2 prediction, while the L2 robust ladder is flat and large (0.26, 0.22, 0.28), with Opus the most scaffold-sensitive Anthropic model rather than the least. The full ladder analysis across all three slices, and the two interpretations of the L2 finding that the data cannot adjudicate between, are in Appendix~\ref{app:tier_sensitivity}.

\section{Discussion}

    \textit{Scaffold choice changes what an evaluation measures.} The headline answer to RQ1 is unambiguous: holding model, tasks, and conditions fixed, scaffold choice alone moves measured GAIA accuracy by as much as 28 points within a single model, the spread between Opus L2's best scaffold (s2, 0.84) and worst (s1, 0.56) under the robust slice. H1's pre-registered threshold of a 10 percentage-point gap for at least one model at Level 1 is cleared by Haiku under all three slices (Section~\ref{sec:scaffold_gap_stat}), and the L2 gaps are larger still for most models. A reader comparing two models evaluated under different scaffolds could easily attribute to the model a difference that belongs to the agent loop. This is the elicitation confound named by \citet{kwa2026measuring} made concrete: the magnitude observed here is in the same range as the roughly 30 percentage-point single-model elicitation gap reported by \citet{pimpale2025} on SWE-Bench Verified, suggesting the phenomenon is not benchmark-specific.

    \textit{Scaffold sensitivity does not diminish with capability.} H2 predicted that more capable models would be less scaffold-sensitive, and the evidence splits. The inferential test rejects equal scaffold effects across models in every slice (Section~\ref{sec:h2_inferential}), so sensitivity is genuinely model-dependent, but the direction contradicts the prediction: most non-Opus interaction coefficients are negative under the robust slice, and the descriptive tier ladder (Section~\ref{sec:tier_sensitivity}) is monotonic in H2's direction only at L1 under that slice. At L2 the robust ladder is flat and large, with Opus the most scaffold-sensitive Anthropic model. Appendix~\ref{app:tier_sensitivity} offers two interpretations the data cannot adjudicate between. Under either, the consequence for evaluation methodology is the same: the elicitation gap is not guaranteed to shrink as models improve, and within the Anthropic family at the harder level it grows.

    \textit{Family, not tier, conditions the multi-agent advantage.} H3 predicted s2 over s1 at L2 but not L1, on the intuition that decomposition pays off when tasks are hard. At L1, no model shows an s2 advantage with a CI above zero under the primary slice; under the robust slice the lone exception is Sonnet (+0.13, CI [0.03, 0.23]). The L2 half held only within the Anthropic family, with robust-slice advantages whose CIs sit strictly above zero, while Gemini and GPT-5 show small advantages indistinguishable from zero. The conditioning variable is model family rather than capability tier: Haiku and Opus sit at opposite ends of the Anthropic ladder yet both benefit, while two frontier models from other providers do not. The behavioral signatures (Section~\ref{sec:rq3}) give this a mechanism-shaped outline without establishing a mechanism: ReAct makes the most tool calls, runs longest, and recovers least often from in-trajectory tool errors at L2, where mid-trajectory errors actually occur. Two candidate explanations are plausible but untested: an error under s2 passes through a rater role positioned to flag and re-route it, and an error under s3 lands inside an explicit plan that may survive the failed step. The data cannot identify the mechanism; the practical reading is that multi-agent structure is not a uniform upgrade, and porting a scaffold finding across model families is not safe.

    \textit{Planning ahead did not help where it was predicted to.} H4 predicted a planner-then-executor advantage on file-reading tasks and is falsified: one supporting cell in ten under the primary slice, none under the robust slice. The two interpretive caveats in Section~\ref{sec:h4}, a Haiku floor on document handling and a Gemini/GPT-5 ceiling on the subset, bound what the falsification means. The honest conclusion is narrower than the hypothesis: on this dataset, with a document subset of sixteen sample\_ids, the only supporting cell does not survive the robust check, and a fair re-test requires a larger file-task pool and models positioned between floor and ceiling on it.

    \textit{Implications for evaluation methodology.} Taken together, the results argue that single-scaffold capability numbers are scaffold-conditional point estimates, not model properties. The interaction structure observed here (scaffold effects that vary by model, invert across difficulty levels, and respond differently across providers) makes a single-cell comparison between two models under two different scaffolds close to uninterpretable as a model comparison. Cost sharpens the point into a procurement decision: Gemini s3 is the cheapest cell at both levels and the most accurate at L2 under all three slices, the within-Anthropic cost dispersion at L2 makes the scaffold choice a budget decision as much as an accuracy decision, and an evaluator with a fixed budget who picks a poor scaffold measures less capability while paying more per unit of measured capability. The actionable version is cheap: running even one alternative scaffold over a benchmark exposes whether the headline number is scaffold-fragile, and fragility is largest precisely where the stakes are highest, on capable models attempting hard tasks.

\section{Limitations}\label{sec:limitations}

    \textit{Tool-surface confound in s1 contrasts.} S1 ran as the unmodified \texttt{inspect\_evals} solver, whose tool surface differs from S2 and S3: a stateful \texttt{web\_browser} in place of \texttt{web\_search}, and no text editor (Section~\ref{subsec:scaffolds}). Contrasts involving s1, including the H3 test, therefore measure loop structure and tool surface jointly; the s2-versus-s3 contrast is the clean structural comparison.
    
    \textit{Log-derived cost versus billed cost.} Every cost figure in this paper is \emph{log-derived}: token counts reported in the Inspect log multiplied by published per-token rates. Four sources of drift separate it from the invoice. Debug and pilot runs hit the bill but are absent from the validation corpus; provider-side hosted tool execution (Anthropic's \texttt{web\_fetch}, Gemini's \texttt{google\_search} grounding) does not always surface in the per-attempt usage fields; cache-read and cache-write accounting conventions are provider-specific and not always aligned with billing; and input-rejected requests (the \texttt{provider\_serialization\_bug}, PDF page-limit, and \texttt{prompt\_too\_long} patterns) may appear on the bill while the log records them as zero-token errors. These drifts are partially offsetting and we did not reconcile them quantitatively. The cost-per-correct headline (Section~\ref{sec:cost_per_correct}) should be read as a directional indicator of comparative cost efficiency, not a procurement quote.

    \textit{Provider-side errors and the SDK patch.} A provider-side incident affected the Anthropic family for the full validation run: Anthropic's API returned a \texttt{web\_fetch\_tool\_result.error\_code} value the published SDK could not deserialize, producing an HTTP 400 on the affected request (patch in Appendix~\ref{app:anthropic_sdk_patch}). The patch, applied for attempt 3, eliminated the deserialization failure but not the provider's underlying return value, and 314 records persisted. Affected requests consumed input tokens before erroring, contributing cost without contributing to the success denominator, with impact concentrated on Opus and Sonnet s2/s3 at L2 (Appendix~\ref{app:provider_bug_summary}); Gemini and GPT-5 were unaffected. Our analytic response is the \emph{robust} slice, reported alongside primary throughout. Where it changes the interpretation we note it: the L2 robust scaffold gap expands the within-Anthropic Sonnet and Opus gaps substantially (Section~\ref{sec:scaffold_gap_stat}) and similarly compresses the cost-per-correct gap (Section~\ref{sec:cost_per_correct}). Readers should treat primary-slice Anthropic L2 numbers as conservative on accuracy and inflated on cost, with the robust slice as the corrected reading.

    \textit{Credit-exhaustion exclusions.} Credit exhaustion mid-flight on the Anthropic family during attempts 2 and 3 terminated dispatch on a subset of units before the provider returned a result; the 156 quarantined \texttt{.eval} files and resume-sweep mechanics are documented in Appendix~\ref{app:incomplete_cells}. The handling is conservative: a dispatch-time rejection never reaches the model, so quarantined records carried no usage and hence no cost or behavioral signal, and their removal does not bias any per-cell statistic. The asymmetry it creates is in retry budget, not the analysis pool: cross-provider models received an honest retry at any failed dispatch, while Anthropic failures from credit exhaustion onward received none, so Anthropic terminal-failure counts at attempts 2 and 3 were not re-tried under matched resource conditions. The analyses operate on units rather than retries, so this does not directly enter the headline statistics; the recovery-from-failure analysis treats each attempt as an independent observation, the conservative reading.

    \textit{Extended-thinking token retention.}\phantomsection\label{sec:limitation_thinking} Reasoning was disabled at the model configuration level for every cell (Section~\ref{subsec:models}), so the analysis intentionally excludes extended-thinking signal; the run's logging configuration, however, also did not retain incidental thinking output. This surfaced during content-filter inspection (Appendix~\ref{sec:content_filter}): a blocked response's trace contains pre-block tool calls and completed messages but not the in-progress reasoning, so the trigger of a block cannot be identified from the trace. The implication for the present analysis is narrow, since the 28 content-filter records do not affect the scaffold, cost, or behavioral analyses. The implication for follow-up work is larger: any analysis of reasoning patterns under scaffold variation requires thinking-stream retention, which the logging configuration of this run did not provide.

    \textit{Validation-set contamination risk for L1.}\phantomsection\label{sec:limitation_contamination} GAIA's validation set is public; the test set is held out behind a leaderboard submission. The high L1 accuracies here (top cells at 84\% to 85\%) may partially reflect memorization rather than capability. One indirect control: if L1 were dominated by memorization, accuracy should track training compute rather than scaffold variation, yet the scaffold gap is similar at both levels for most models (Section~\ref{sec:scaffold_gap_stat}), suggesting scaffolds exert real variation at L1. The concern attaches more strongly to the L1 cost-per-correct rankings than to the scaffold-effect findings, since a memorization-inflated accuracy denominator would underestimate true cost.

\section{Conclusion and future work}

    This study executed a pre-registered controlled comparison of three agent scaffolds across five models on GAIA Levels 1 and 2, holding tasks and conditions fixed. H1 is supported: scaffold variation produces gaps exceeding the 10 percentage-point threshold, with the largest within-model spread at 28 points (Opus L2, robust slice). H2 is rejected in direction: scaffold effects vary significantly by model in every slice, but the most capable Anthropic model gains the most from structured scaffolds at the harder level. H3 is conditionally supported: the multi-agent scaffold outperforms ReAct at L2 within the Anthropic family but not for the cross-provider models. H4 is falsified: the planner-then-executor advantage on file-reading tasks does not survive the robust check in any cell. Beyond the hypothesis tests, structured scaffolds make fewer tool calls, finish faster, and recover more often from mid-trajectory errors at L2; Gemini s3 is the cheapest cell at both levels and the most accurate at L2 under all three slices.

    The central result for evaluation methodology: the elicitation gap named by \citet{kwa2026measuring} is large, measurable under controlled conditions, and not guaranteed to shrink as models improve. Capability numbers produced under a single scaffold are scaffold-conditional estimates, and comparisons across differently-scaffolded evaluations confound model capability with agent-loop fit.

    Three extensions follow directly: held-out test-set submission, one run per (model, scaffold) cell against the GAIA leaderboard, as a contamination-free check on the L1 findings (Section~\ref{sec:limitation_contamination}); a logging configuration that retains reasoning tokens, enabling analysis of in-trajectory adaptation under scaffold variation (Section~\ref{sec:limitation_thinking}); and, from the family-conditional H3 result and the falsified H4, a wider model panel to test whether the multi-agent advantage tracks provider boundaries, and a larger file-task pool with models between floor and ceiling for a fair planner-executor re-test.

\bibliography{main}
\bibliographystyle{tmlr}

\appendix
\section{Inspect AI Gemini provider patch}\label{app:gemini_patch}

    When using Gemini 3 Pro with both Gemini's native \texttt{web\_search} (which resolves to \texttt{google\_search}) and function tools (\texttt{bash}, \texttt{python}, \texttt{text\_editor}), Inspect AI 0.3.217 raised a pre-flight \texttt{ValueError} reading "Gemini does not yet support native web search or code execution concurrently with other tools." The underlying API limitation was lifted by Google in March 2026 with the Gemini 3 series via the \texttt{include\_server\_side\_tool\_invocations} flag; as of Inspect 0.3.219, support for the flag had not been added.
    
    Two edits to \texttt{inspect\_ai/model/\_providers/google.py} resolve this.
    
        \subsection{Edit 1: ToolConfig with the flag}
        
            In the \texttt{generate} method of \texttt{GoogleGenAIAPI} (around line 316), construct a \texttt{ToolConfig} with \texttt{include\_server\_side\_tool\_invocations=True} when both native and function tools are present.
            
            Before:
            \begin{verbatim}
            gemini_tool_config = (
                chat_tool_config(tool_choice)
                if not has_native_tools and len(tools) > 0
                else None
            )
            \end{verbatim}
            
            After:
            \begin{verbatim}
            if not has_native_tools and len(tools) > 0:
                gemini_tool_config = chat_tool_config(tool_choice)
            elif has_native_tools and len(tools) > 0:
                from google.genai.types import ToolConfig
                gemini_tool_config = ToolConfig(
                    include_server_side_tool_invocations=True
                )
            else:
                gemini_tool_config = None
            \end{verbatim}
        
        \subsection{Edit 2: Remove the pre-flight ValueError}
        
            In the \texttt{chat\_tools} method (around line 891), remove the pre-flight guard and append function declarations alongside native tools.
            
            Before:
            \begin{verbatim}
            if google_search or code_execution:
                if function_declarations:
                    raise ValueError(
                        "Gemini does not yet support native web search "
                        "or code execution concurrently with other tools."
                    )
                native_tools: ToolListUnion = []
                if google_search:
                    native_tools.append(Tool(google_search=google_search))
                if code_execution:
                    native_tools.append(Tool(code_execution=code_execution))
                return (True, native_tools)
            \end{verbatim}
            
            After:
            \begin{verbatim}
            if google_search or code_execution:
                native_tools: ToolListUnion = []
                if google_search:
                    native_tools.append(Tool(google_search=google_search))
                if code_execution:
                    native_tools.append(Tool(code_execution=code_execution))
                if function_declarations:
                    native_tools.append(
                        Tool(function_declarations=function_declarations)
                    )
                return (True, native_tools)
            \end{verbatim}
        
        \subsection{Verification}
        
            The patch was verified by running Gemini 3.1 Pro on GAIA L1 sample \texttt{8e867cd7-cff9-4e6c-867a-ff5ddc2550be} with \texttt{web\_search + bash + python + text\_editor} active; pre-patch the run errored, post-patch it ran to completion and scored accuracy 1.000.
        
        \subsection{Durability}
        
            The patch is an environment workaround, not a contribution to Inspect. It survives \texttt{pip install --upgrade google-genai}, normal venv use, and IDE restarts. It is overwritten by \texttt{pip install --upgrade inspect-ai}, force-reinstalls, or venv recreation, and must be re-applied after any such operation. The patched provider file is included in the project repository\footnote{Anonymized repository: \url{https://anonymous.4open.science/r/gaia-scaffold-comparison-ECBF/}}.

\section{Anthropic SDK patch for \texttt{web\_fetch\_tool\_result} error code}\label{app:anthropic_sdk_patch}

    Midway through the validation run, Anthropic's API began returning a \texttt{web\_fetch\_tool\_result} structure whose \texttt{error\_code} field held the value \texttt{"url\_not\_in\_prior\_context"}. This value is not in the published Anthropic Python SDK's \texttt{Literal} alias for that field, so the SDK's Pydantic model raised a validation error during deserialization. Because the conversation continued and the now-malformed structure was echoed back to Anthropic on the subsequent request, the API responded with an HTTP 400, terminating the attempt. The pattern affected the Anthropic family at attempt-3 cells where Inspect's compaction handler had removed earlier URL-bearing tool calls from context (Appendix~\ref{app:provider_bug_summary}).
    
    Two edits to the Anthropic Python SDK accept the unmapped value and let the SDK deserialize the response cleanly. The patch does not suppress the underlying error: the model still surfaces the \texttt{web\_fetch} failure as a normal tool error. The patch simply prevents that error from cascading into a Pydantic crash and a subsequent HTTP 400 on the next request.
    
        \subsection{Edit 1: Stable types}
        
            In \texttt{anthropic/types/web\_fetch\_tool\_result\_error\_code.py}, extend the \texttt{Literal} alias to include the new value.
            
            Before:
            \begin{verbatim}
            WebFetchToolResultErrorCode: TypeAlias = Literal[
                "invalid_tool_input",
                "url_too_long",
                "url_not_allowed",
                "url_not_accessible",
                "unsupported_content_type",
                "too_many_requests",
                "max_uses_exceeded",
                "unavailable",
            ]
            \end{verbatim}
            
            After:
            \begin{verbatim}
            WebFetchToolResultErrorCode: TypeAlias = Literal[
                "invalid_tool_input",
                "url_too_long",
                "url_not_allowed",
                "url_not_accessible",
                "unsupported_content_type",
                "too_many_requests",
                "max_uses_exceeded",
                "unavailable",
                "url_not_in_prior_context",
            ]
            \end{verbatim}
        
        \subsection{Edit 2: Beta types}
        
            The same change is applied to \texttt{anthropic/types/beta/beta\_web\_fetch\_tool\_result\_error\_code.py}. The beta surface duplicates the stable alias under the \texttt{BetaWebFetchToolResultErrorCode} name; Inspect routes some Anthropic requests through the beta API surface, so both files must be patched for the fix to cover every request path through the SDK.
            
            Before:
            \begin{verbatim}
            BetaWebFetchToolResultErrorCode: TypeAlias = Literal[
                "invalid_tool_input",
                "url_too_long",
                "url_not_allowed",
                "url_not_accessible",
                "unsupported_content_type",
                "too_many_requests",
                "max_uses_exceeded",
                "unavailable",
            ]
            \end{verbatim}
            
            After:
            \begin{verbatim}
            BetaWebFetchToolResultErrorCode: TypeAlias = Literal[
                "invalid_tool_input",
                "url_too_long",
                "url_not_allowed",
                "url_not_accessible",
                "unsupported_content_type",
                "too_many_requests",
                "max_uses_exceeded",
                "unavailable",
                "url_not_in_prior_context",
            ]
            \end{verbatim}
        
        \subsection{Verification}
        
            The patch was verified during the investigation that produced it. Four samples from the original attempt 2 bloat cohort which had terminated with the Pydantic validation error were re-run with the patch in place. Pre-patch, all four terminated on the SDK-side deserialization error followed by an HTTP 400 on the next request. Post-patch, the SDK accepted the response, the \texttt{web\_fetch} error surfaced to the agent loop as a normal tool error, and the attempts either continued or terminated with a clean error in the trace. The patch was then applied for attempt 3 of the full validation run; the 314 records that nonetheless ended up flagged with \texttt{provider\_serialization\_bug} (Appendix~\ref{app:provider_bug_summary}) are records where Anthropic's API returned the unmapped value but the patched SDK accepted it and the agent loop subsequently terminated on the surfaced tool error, rather than crashing on deserialization.
        
        \subsection{Durability}
        
            The patch is an environment workaround, not a contribution to the Anthropic SDK. It survives normal venv use and IDE restarts. It is overwritten by \texttt{pip install --upgrade anthropic}, \texttt{pip install --force-reinstall anthropic}, or venv recreation, and must be re-applied after any such operation. The patched SDK files are included in the project repository\footnote{Anonymized repository: \url{https://anonymous.4open.science/r/gaia-scaffold-comparison-ECBF/}}.

\section{Scaffold prompts}\label{app:scaffold_prompts}

    This appendix contains the full role and phase prompts for the S2 (PAR) and S3 (Planner-then-Executor) scaffolds. S1 (ReAct) uses the default \texttt{inspect\_evals} GAIA prompt and is not reproduced here.
    
    \subsection{S2: Planner-Actor-Rater (PAR)}\label{app:par_prompts}
    
        \textit{Planner prompt:}
        
        \begin{quote}
        You are the Advisor. Read the question and any prior actor and rater turns in the conversation. Produce a short bulleted plan of next steps for the Actor to execute. Do not attempt to solve the task yourself. Do not produce an answer. Do not call tools. If a prior plan exists and the rater said CONTINUE, revise the plan based on the rater's feedback and what the actor has already tried.
        \end{quote}
        
        \textit{Actor prompt:}
        
        \begin{quote}
        You are the Actor. You will find the answer to the question following the most recent bulleted plan. You will disregard any comments beginning with CONTINUE. Note any prior steps that successfully gathered information; build on them rather than redoing them. Do not repeat actions that previously failed. You have access to (bash, python, web\_search, text\_editor, submit\_answer) use them as you see fit to find the answer, when you have your final answer you must use submit\_answer. Your answer will be in the format provided in the original question. Do not use markdown formatting (bold, italics, code blocks, bullet points), do not prefix with "Answer:" or similar labels, do not add explanation. Return only the bare value matching the format specified in the question.
        \end{quote}
        
        \textit{Rater prompt:}
        
        \begin{quote}
        You are an evaluator. Review the most recent answer. Decide whether it correctly addresses the question and matches the format required. Call submit\_verdict with done=True if the answer is correct and well-formatted, or done=False if more work is needed. When done=False, include a short reason explaining what is wrong.
        \end{quote}
    
    \subsection{S3: Planner-then-Executor}\label{app:pte_prompts}
    
        \textit{Planner prompt:}
        
        \begin{quote}
        You are the planner, your job is to evaluate the question you receive and create a detailed bulleted plan for the executor. You will restate the format requirements from the question so the executor does not miss them. You do not have tools, do not execute the plan, and do not answer the question. You will only provide the plan.
        \end{quote}
        
        \textit{Executor prompt:}
        
        \begin{quote}
        You are the executor. You will receive a detailed plan from the planner. You will follow this plan to find an answer to the provided question. Your response will be in the format provided in the question. Return only the bare value matching the format specified in the question.
        \end{quote}

\section{Incomplete cells and failure analysis}\label{app:incomplete_cells}\label{sec:provider_bug}\label{sec:prompt_too_long}\label{sec:content_filter}\label{sec:pdf_and_parsing}

    Five failure modes account for every non-completed attempt-unit in the validation run. First, two Anthropic credit-exhaustion events produced 1{,}446 provider-rejected dispatches with zero tokens and zero working time, quarantined in 156 \texttt{.eval} files before analysis; the affected cells received clean retries in the resume sweep. The counts are from the pre-cleaning logs; the released log tree is the post-quarantine state. Second, 314 records carry the \texttt{provider\_serialization\_bug} flag: Anthropic's API returned a \texttt{web\_fetch\_tool\_result} error code the published SDK could not deserialize (Appendix~\ref{app:anthropic_sdk_patch}); none received a clean retry, motivating the robust slice. Per-cell concentrations are in Appendix~\ref{app:provider_bug_summary}. Third, 39 records (33 unique units) terminated with a \texttt{prompt is too long} error, 27 of 33 in Haiku s2 and recurring on a small set of GAIA questions, indicating systematic context pressure rather than stochastic variation; retry threading is in Appendix~\ref{app:prompt_too_long_threading}. Fourth, 28 records terminated on a provider content filter, corresponding to two underlying GAIA questions (Appendix~\ref{app:content_filter_inspection}); the logging limitation this surfaced is discussed in Section~\ref{sec:limitation_thinking}. Fifth, ten records terminated on provider-side input rejections independent of scaffold or capability: nine on the Anthropic 100-PDF-page limit and one on a null-byte read failure.

    Of the 30 (model, scaffold, level) cells in the validation design, 15 completed every attempt-unit and 15 are partial. All partial cells belong to the Anthropic family; every Gemini and GPT-5 cell completed at 100\%. Table~\ref{tab:incomplete_cells} lists each partial cell with its completion rate and the per-bucket breakdown of errored units using the failure buckets defined above. Bucket counts sum exactly to the errored-unit count in every row. Credit-exhausted dispatches do not appear here. Two Anthropic credit-exhaustion events during the validation run (a partial drain at the end of attempt 2 affecting Opus, and a complete drain of all Anthropic attempt 3 cells before credits were reloaded) produced 1{,}446 sample-level API rejections across 21 (model, scaffold, attempt, level) cells. The rejections concentrate in 156 \texttt{.eval} files in which every contained sample matched the same provider response (\texttt{credit balance is too low}); no \texttt{.eval} file mixed valid completions with credit-exhausted samples. All 156 files were quarantined before any downstream computation, and the May 24 resume sweep produced clean retries for the affected cells, so the cells are not lost to the analysis, only the rejected dispatches. Across all 30 cells, 5{,}907 of the 6{,}255 designed attempt-units completed without an Inspect-level error.

    \begin{table}[h]
        \caption{Partial cells (completion rate below 1.00) with errored-unit counts by failure bucket. n\_att is designed attempt-units (159 at L1, 258 at L2); n\_comp is units completing without an Inspect-level error; n\_err is n\_att minus n\_comp. Bucket abbreviations: ptl = \texttt{prompt\_too\_long}, cf = content filter, pdf = PDF page-limit, psb = \texttt{provider\_serialization\_bug}.}
    \label{tab:incomplete_cells}
    \begin{center}
    \small
    \begin{tabular}{lllrrrrrrrr}
        \multicolumn{1}{c}{\bf model} & \multicolumn{1}{c}{\bf scaffold} & \multicolumn{1}{c}{\bf level} & \multicolumn{1}{c}{\bf n\_att} & \multicolumn{1}{c}{\bf n\_comp} & \multicolumn{1}{c}{\bf n\_err} & \multicolumn{1}{c}{\bf comp.\ rate} & \multicolumn{1}{c}{\bf ptl} & \multicolumn{1}{c}{\bf cf} & \multicolumn{1}{c}{\bf pdf} & \multicolumn{1}{c}{\bf psb} \\
        \hline \\
        haiku  & s2 & L1 & 159 & 150 &  9 & 94.3\% & 1 & 0 & 0 &  8 \\
        haiku  & s3 & L1 & 159 & 154 &  5 & 96.9\% & 0 & 0 & 0 &  5 \\
        opus   & s2 & L1 & 159 & 150 &  9 & 94.3\% & 0 & 0 & 0 &  9 \\
        opus   & s3 & L1 & 159 & 134 & 25 & 84.3\% & 0 & 0 & 0 & 25 \\
        sonnet & s1 & L1 & 159 & 158 &  1 & 99.4\% & 0 & 1 & 0 &  0 \\
        sonnet & s2 & L1 & 159 & 147 & 12 & 92.5\% & 0 & 0 & 0 & 12 \\
        sonnet & s3 & L1 & 159 & 145 & 14 & 91.2\% & 0 & 1 & 0 & 13 \\
        haiku  & s1 & L2 & 258 & 257 &  1 & 99.6\% & 0 & 1 & 0 &  0 \\
        haiku  & s2 & L2 & 258 & 208 & 50 & 80.6\% & 5 & 0 & 4 & 41 \\
        haiku  & s3 & L2 & 258 & 228 & 30 & 88.4\% & 4 & 3 & 2 & 21 \\
        opus   & s1 & L2 & 258 & 257 &  1 & 99.6\% & 0 & 1 & 0 &  0 \\
        opus   & s2 & L2 & 258 & 212 & 46 & 82.2\% & 1 & 2 & 0 & 43 \\
        opus   & s3 & L2 & 258 & 205 & 53 & 79.5\% & 0 & 0 & 0 & 53 \\
        sonnet & s2 & L2 & 258 & 212 & 46 & 82.2\% & 0 & 3 & 0 & 43 \\
        sonnet & s3 & L2 & 258 & 212 & 46 & 82.2\% & 2 & 3 & 0 & 41 \\
    \end{tabular}
    \end{center}
    \end{table}

\section{Full per-cell outcomes}\label{app:per_cell_outcomes_full}

    Table~\ref{tab:per_cell_outcomes_full} reports per-cell outcomes for all five models: attempt counts, completion rate, the failure-mode bucket counts that decompose the difference between \texttt{n\_att} and \texttt{correct}, and both accuracy definitions used in Section~\ref{sec:per_cell_outcomes}. Every attempt-unit falls into exactly one bucket: \texttt{correct} (completed and scored as correct by the GAIA scorer), \texttt{wrong} (completed and scored as incorrect, including time-limit truncations that submitted no answer), \texttt{p2l} (terminated with a \texttt{prompt is too long} provider error, Appendix~\ref{sec:prompt_too_long}), \texttt{cf} (terminated with a content-filter block, Appendix~\ref{sec:content_filter}), \texttt{pdf} (terminated with the Anthropic 100-PDF-page limit, Appendix~\ref{sec:pdf_and_parsing}), \texttt{bug} (carried the \texttt{provider\_serialization\_bug} flag, Appendix~\ref{sec:provider_bug}), and \texttt{other} (any other Inspect-level error not matching the above signatures). Bucket counts sum to \texttt{n\_att} on every row, providing a complete accounting of where every dispatched attempt ended up.

    \begin{table}[t]
        \caption{Full per-cell outcomes for the validation run. Rows are sorted by (level, model, scaffold). No attempt-unit terminated in the \texttt{other} bucket, so the column is omitted. comp.\ rate is the fraction of attempt-units that did not terminate with an Inspect-level error; acc.\ overall divides \texttt{correct} by \texttt{n\_att}; acc.\ comp.\ divides \texttt{correct} by the non-errored count.}
    \label{tab:per_cell_outcomes_full}
    \begin{center}
    \footnotesize \setlength{\tabcolsep}{3pt}
    \begin{tabular}{lllrrrrrrrrrr}
    \multicolumn{1}{c}{\bf level} & \multicolumn{1}{c}{\bf model} & \multicolumn{1}{c}{\bf scaffold} & \multicolumn{1}{c}{\bf n\_att} & \multicolumn{1}{c}{\bf comp.\ rate} & \multicolumn{1}{c}{\bf correct} & \multicolumn{1}{c}{\bf wrong} & \multicolumn{1}{c}{\bf p2l} & \multicolumn{1}{c}{\bf cf} & \multicolumn{1}{c}{\bf pdf} & \multicolumn{1}{c}{\bf bug} & \multicolumn{1}{c}{\bf acc.\ overall} & \multicolumn{1}{c}{\bf acc.\ comp.} \\
        \hline \\
        L1 & haiku & s1 & 159 & 1.00 & 73 & 86 & 0 & 0 & 0 & 0 & 0.459 & 0.459 \\
        L1 & haiku & s2 & 159 & 0.94 & 73 & 77 & 1 & 0 & 0 & 8 & 0.459 & 0.487 \\
        L1 & haiku & s3 & 159 & 0.97 & 36 & 118 & 0 & 0 & 0 & 5 & 0.226 & 0.234 \\
        L1 & sonnet & s1 & 159 & 0.99 & 105 & 53 & 0 & 1 & 0 & 0 & 0.660 & 0.665 \\
        L1 & sonnet & s2 & 159 & 0.92 & 116 & 31 & 0 & 0 & 0 & 12 & 0.730 & 0.789 \\
        L1 & sonnet & s3 & 159 & 0.91 & 116 & 29 & 0 & 1 & 0 & 13 & 0.730 & 0.800 \\
        L1 & opus & s1 & 159 & 1.00 & 132 & 27 & 0 & 0 & 0 & 0 & 0.830 & 0.830 \\
        L1 & opus & s2 & 159 & 0.94 & 121 & 29 & 0 & 0 & 0 & 9 & 0.761 & 0.807 \\
        L1 & opus & s3 & 159 & 0.84 & 112 & 22 & 0 & 0 & 0 & 25 & 0.704 & 0.836 \\
        L1 & gemini-3.1-pro & s1 & 159 & 1.00 & 125 & 34 & 0 & 0 & 0 & 0 & 0.786 & 0.786 \\
        L1 & gemini-3.1-pro & s2 & 159 & 1.00 & 132 & 27 & 0 & 0 & 0 & 0 & 0.830 & 0.830 \\
        L1 & gemini-3.1-pro & s3 & 159 & 1.00 & 134 & 25 & 0 & 0 & 0 & 0 & 0.843 & 0.843 \\
        L1 & gpt-5 & s1 & 159 & 1.00 & 129 & 30 & 0 & 0 & 0 & 0 & 0.811 & 0.811 \\
        L1 & gpt-5 & s2 & 159 & 1.00 & 134 & 25 & 0 & 0 & 0 & 0 & 0.843 & 0.843 \\
        L1 & gpt-5 & s3 & 159 & 1.00 & 135 & 24 & 0 & 0 & 0 & 0 & 0.849 & 0.849 \\
        L2 & haiku & s1 & 258 & 1.00 & 81 & 176 & 0 & 1 & 0 & 0 & 0.314 & 0.315 \\
        L2 & haiku & s2 & 258 & 0.81 & 110 & 98 & 5 & 0 & 4 & 41 & 0.426 & 0.529 \\
        L2 & haiku & s3 & 258 & 0.88 & 58 & 170 & 4 & 3 & 2 & 21 & 0.225 & 0.254 \\
        L2 & sonnet & s1 & 258 & 1.00 & 135 & 123 & 0 & 0 & 0 & 0 & 0.523 & 0.523 \\
        L2 & sonnet & s2 & 258 & 0.82 & 160 & 52 & 0 & 3 & 0 & 43 & 0.620 & 0.755 \\
        L2 & sonnet & s3 & 258 & 0.82 & 154 & 58 & 2 & 3 & 0 & 41 & 0.597 & 0.726 \\
        L2 & opus & s1 & 258 & 1.00 & 144 & 113 & 0 & 1 & 0 & 0 & 0.558 & 0.560 \\
        L2 & opus & s2 & 258 & 0.82 & 180 & 32 & 1 & 2 & 0 & 43 & 0.698 & 0.849 \\
        L2 & opus & s3 & 258 & 0.79 & 169 & 36 & 0 & 0 & 0 & 53 & 0.655 & 0.824 \\
        L2 & gemini-3.1-pro & s1 & 258 & 1.00 & 179 & 79 & 0 & 0 & 0 & 0 & 0.694 & 0.694 \\
        L2 & gemini-3.1-pro & s2 & 258 & 1.00 & 194 & 64 & 0 & 0 & 0 & 0 & 0.752 & 0.752 \\
        L2 & gemini-3.1-pro & s3 & 258 & 1.00 & 222 & 36 & 0 & 0 & 0 & 0 & 0.860 & 0.860 \\
        L2 & gpt-5 & s1 & 258 & 1.00 & 197 & 61 & 0 & 0 & 0 & 0 & 0.764 & 0.764 \\
        L2 & gpt-5 & s2 & 258 & 1.00 & 201 & 57 & 0 & 0 & 0 & 0 & 0.779 & 0.779 \\
        L2 & gpt-5 & s3 & 258 & 1.00 & 212 & 46 & 0 & 0 & 0 & 0 & 0.822 & 0.822 \\
    \end{tabular}
    \end{center}
    \end{table}

\FloatBarrier

\section{Full bootstrap accuracy}\label{app:bootstrap_accuracy_full}

\begin{figure}
        \centering
        \includegraphics[width=0.99\linewidth]{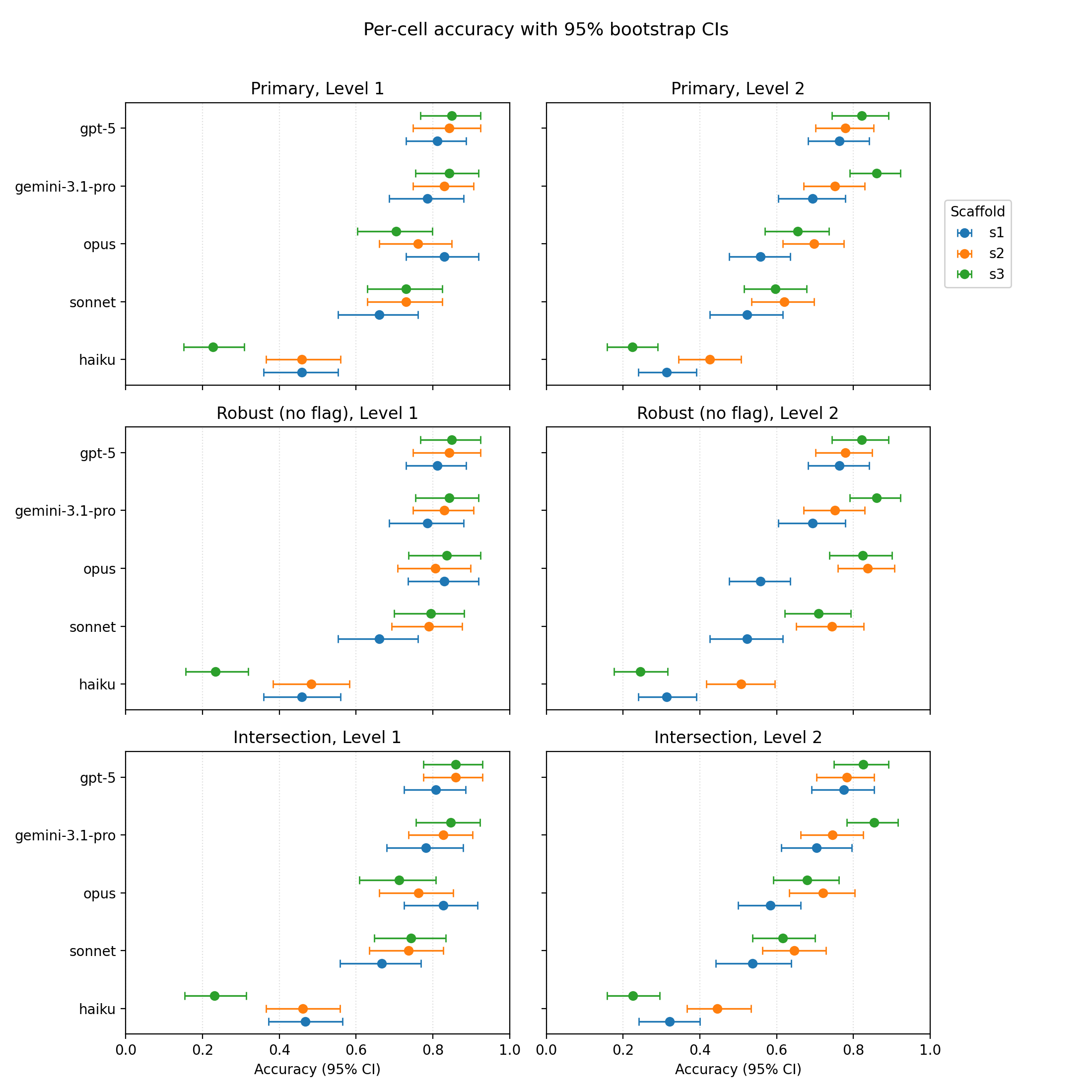}
        \caption{Per-cell accuracy with 95\% bootstrap CIs, by slice (rows) and GAIA level (columns); within each panel, one dot per scaffold.}
        \label{fig:per_cell_accuracy}
    \end{figure}

    Table~\ref{tab:bootstrap_accuracy_full} reports the bootstrap accuracy point estimate and 95\% CI for every (model, scaffold, level) cell across the primary, robust (no flag), and intersection slices. The same data is visualized in Figure~\ref{fig:per_cell_accuracy}. Each CI is computed from 5{,}000 sample\_id-clustered bootstrap resamples with seed 20260526. The same seed is used across the three slices so resample draws are identical at the level of sample\_id picks; differences between slices reflect only the input pool. The cluster count per cell is 53 sample\_ids at Level 1 and 86 at Level 2 in the primary slice; the intersection slice retains 52 of 53 L1 sample\_ids and 80 of 86 L2 sample\_ids, while the robust slice retains all sample\_ids but drops attempt-units carrying the 
    \texttt{provider\_serialization\_bug} flag.

    \begin{table}[t]
        \caption{Full bootstrap accuracy for the validation run. Rows are sorted by (level, model, scaffold). Each cell reports point accuracy and the 95\% bootstrap CI. The primary slice includes all attempt-units; the robust (no flag) slice excludes units carrying the \texttt{provider\_serialization\_bug} flag; the intersection slice retains only sample\_ids that completed in every (model, scaffold) cell at the corresponding level.}
    \label{tab:bootstrap_accuracy_full}
    \begin{center}
    \small
    \begin{tabular}{llllll}
    \multicolumn{1}{c}{\bf level} & \multicolumn{1}{c}{\bf model} & \multicolumn{1}{c}{\bf scaffold} & \multicolumn{1}{c}{\bf primary [95\% CI]} & \multicolumn{1}{c}{\bf robust [95\% CI]} & \multicolumn{1}{c}{\bf intersection [95\% CI]} \\
        \hline \\
        L1 & haiku & s1 & 0.459 [0.358, 0.553] & 0.459 [0.358, 0.560] & 0.468 [0.372, 0.564] \\
        L1 & haiku & s2 & 0.459 [0.365, 0.560] & 0.483 [0.383, 0.583] & 0.462 [0.365, 0.558] \\
        L1 & haiku & s3 & 0.226 [0.151, 0.308] & 0.234 [0.156, 0.318] & 0.231 [0.154, 0.314] \\
        L1 & sonnet & s1 & 0.660 [0.553, 0.761] & 0.660 [0.553, 0.761] & 0.667 [0.558, 0.769] \\
        L1 & sonnet & s2 & 0.730 [0.629, 0.824] & 0.789 [0.692, 0.877] & 0.737 [0.635, 0.827] \\
        L1 & sonnet & s3 & 0.730 [0.629, 0.824] & 0.795 [0.699, 0.882] & 0.744 [0.647, 0.833] \\
        L1 & opus & s1 & 0.830 [0.730, 0.918] & 0.830 [0.736, 0.918] & 0.827 [0.724, 0.917] \\
        L1 & opus & s2 & 0.761 [0.660, 0.849] & 0.807 [0.708, 0.898] & 0.763 [0.660, 0.853] \\
        L1 & opus & s3 & 0.704 [0.604, 0.799] & 0.836 [0.736, 0.924] & 0.712 [0.609, 0.808] \\
        L1 & gemini-3.1-pro & s1 & 0.786 [0.686, 0.881] & 0.786 [0.686, 0.881] & 0.782 [0.679, 0.878] \\
        L1 & gemini-3.1-pro & s2 & 0.830 [0.748, 0.906] & 0.830 [0.748, 0.906] & 0.827 [0.737, 0.904] \\
        L1 & gemini-3.1-pro & s3 & 0.843 [0.755, 0.918] & 0.843 [0.755, 0.918] & 0.846 [0.756, 0.923] \\
        L1 & gpt-5 & s1 & 0.811 [0.730, 0.887] & 0.811 [0.730, 0.887] & 0.808 [0.724, 0.885] \\
        L1 & gpt-5 & s2 & 0.843 [0.748, 0.925] & 0.843 [0.748, 0.925] & 0.859 [0.776, 0.930] \\
        L1 & gpt-5 & s3 & 0.849 [0.767, 0.925] & 0.849 [0.767, 0.925] & 0.859 [0.776, 0.929] \\
        L2 & haiku & s1 & 0.314 [0.240, 0.391] & 0.314 [0.240, 0.391] & 0.321 [0.242, 0.400] \\
        L2 & haiku & s2 & 0.426 [0.345, 0.508] & 0.507 [0.417, 0.596] & 0.446 [0.367, 0.533] \\
        L2 & haiku & s3 & 0.225 [0.159, 0.291] & 0.245 [0.176, 0.316] & 0.225 [0.158, 0.296] \\
        L2 & sonnet & s1 & 0.523 [0.426, 0.616] & 0.523 [0.426, 0.616] & 0.537 [0.442, 0.637] \\
        L2 & sonnet & s2 & 0.620 [0.535, 0.698] & 0.744 [0.651, 0.827] & 0.646 [0.562, 0.729] \\
        L2 & sonnet & s3 & 0.597 [0.516, 0.678] & 0.710 [0.621, 0.794] & 0.617 [0.537, 0.700] \\
        L2 & opus & s1 & 0.558 [0.477, 0.636] & 0.558 [0.477, 0.636] & 0.583 [0.500, 0.662] \\
        L2 & opus & s2 & 0.698 [0.616, 0.775] & 0.837 [0.759, 0.907] & 0.721 [0.633, 0.804] \\
        L2 & opus & s3 & 0.655 [0.570, 0.736] & 0.824 [0.738, 0.900] & 0.679 [0.592, 0.762] \\
        L2 & gemini-3.1-pro & s1 & 0.694 [0.605, 0.779] & 0.694 [0.605, 0.779] & 0.704 [0.613, 0.796] \\
        L2 & gemini-3.1-pro & s2 & 0.752 [0.671, 0.829] & 0.752 [0.671, 0.829] & 0.746 [0.662, 0.825] \\
        L2 & gemini-3.1-pro & s3 & 0.860 [0.791, 0.922] & 0.860 [0.791, 0.922] & 0.854 [0.783, 0.917] \\
        L2 & gpt-5 & s1 & 0.764 [0.682, 0.841] & 0.764 [0.682, 0.841] & 0.775 [0.692, 0.854] \\
        L2 & gpt-5 & s2 & 0.779 [0.702, 0.853] & 0.779 [0.702, 0.849] & 0.783 [0.704, 0.854] \\
        L2 & gpt-5 & s3 & 0.822 [0.744, 0.891] & 0.822 [0.744, 0.891] & 0.825 [0.750, 0.892] \\
    \end{tabular}
    \end{center}
    \end{table}

\FloatBarrier

\section{Full scaffold gap}\label{app:scaffold_gap_full}

    Table~\ref{tab:scaffold_gap_full} reports the per-(model, level) scaffold gap (max minus min accuracy across the three scaffolds) under the three slices, including which scaffold is at the max and min for each slice. The gap point estimate and 95\% CI come from 5{,}000 sample\_id-clustered bootstrap resamples shared across scaffolds within each (model, level) iteration. The same data is visualized in Figure~\ref{fig:scaffold_gap_bars} in the body. Cases where the robust or intersection slice flips which scaffold sits at the max or min are visible by comparing the parenthesized pairs across columns within a row: the L1 opus row, for example, shows (s1, s3) under primary but (s3, s2) under robust, reflecting that removing flagged units changes which scaffold has the highest and lowest accuracy.

    \begin{table}[t]
        \caption{Full scaffold gap for the validation run. Each non-key column reports the gap point estimate and 95\% bootstrap CI followed in parentheses by the scaffold at the max and the scaffold at the min for that slice.}
    \label{tab:scaffold_gap_full}
    \begin{center}
    \small
    \begin{tabular}{lllll}
        \multicolumn{1}{c}{\bf level} & \multicolumn{1}{c}{\bf model} & \multicolumn{1}{c}{\bf primary} & \multicolumn{1}{c}{\bf robust (no flag)} & \multicolumn{1}{c}{\bf intersection} \\
        \hline \\
        L1 & haiku & 0.233 [0.157, 0.358] (s1, s3) & 0.250 [0.168, 0.363] (s2, s3) & 0.237 [0.154, 0.359] (s1, s3) \\
        L1 & sonnet & 0.069 [0.013, 0.182] (s2, s1) & 0.134 [0.056, 0.244] (s3, s1) & 0.077 [0.019, 0.186] (s3, s1) \\
        L1 & opus & 0.126 [0.038, 0.226] (s1, s3) & 0.029 [0.010, 0.102] (s3, s2) & 0.115 [0.032, 0.218] (s1, s3) \\
        L1 & gemini-3.1-pro & 0.057 [0.013, 0.157] (s3, s1) & 0.057 [0.013, 0.157] (s3, s1) & 0.064 [0.019, 0.167] (s3, s1) \\
        L1 & gpt-5 & 0.038 [0.013, 0.119] (s3, s1) & 0.038 [0.013, 0.119] (s3, s1) & 0.051 [0.013, 0.128] (s2, s1) \\
        L2 & haiku & 0.202 [0.120, 0.287] (s2, s3) & 0.262 [0.174, 0.358] (s2, s3) & 0.221 [0.142, 0.304] (s2, s3) \\
        L2 & sonnet & 0.097 [0.023, 0.198] (s2, s1) & 0.221 [0.126, 0.332] (s2, s1) & 0.108 [0.029, 0.208] (s2, s1) \\
        L2 & opus & 0.140 [0.066, 0.221] (s2, s1) & 0.279 [0.213, 0.357] (s2, s1) & 0.137 [0.062, 0.225] (s2, s1) \\
        L2 & gemini-3.1-pro & 0.167 [0.097, 0.240] (s3, s1) & 0.167 [0.097, 0.240] (s3, s1) & 0.150 [0.079, 0.225] (s3, s1) \\
        L2 & gpt-5 & 0.058 [0.019, 0.132] (s3, s1) & 0.058 [0.019, 0.132] (s3, s1) & 0.050 [0.017, 0.129] (s3, s1) \\
    \end{tabular}
    \end{center}
    \end{table}

\FloatBarrier

\section{Pairwise scaffold differences}\label{app:scaffold_gap_pairwise}

    Table~\ref{tab:scaffold_gap_pairwise_full} reports pairwise scaffold-accuracy differences (e.g., \texttt{s1-s2} is accuracy on s1 minus accuracy on s2) for each (model, level), across the three slices. Three pairs per cell are computed: \texttt{s1-s2}, \texttt{s1-s3}, and \texttt{s2-s3}. Differences are reported with their 95\% bootstrap CIs computed from the same sample\_id-clustered resamples that produced the headline gap in Appendix~\ref{app:scaffold_gap_full}; the resamples are shared across scaffolds within each iteration so the pairwise-difference CI accounts for the within-question dependence across scaffold conditions. A CI excluding zero indicates that the two scaffolds produce statistically distinguishable accuracy on the slice; a CI spanning zero indicates the scaffolds are not distinguishable at the 95\% level. Signs are preserved (positive means the first-named scaffold scored higher; negative means the second scored higher) so the direction of any individual pair is unambiguous.

    \begin{table}[t]
        \caption{Pairwise scaffold differences with 95\% bootstrap CIs for the validation run. Each row reports one (model, level, pair); the three columns at right give the point difference and 95\% CI under the primary, robust (no flag), and intersection slices. Positive values indicate the first scaffold of the pair scored higher than the second on that slice.}
    \label{tab:scaffold_gap_pairwise_full}
    \begin{center}
    \small
    \begin{tabular}{llllll}
        \multicolumn{1}{c}{\bf level} & \multicolumn{1}{c}{\bf model} & \multicolumn{1}{c}{\bf pair} & \multicolumn{1}{c}{\bf primary [95\% CI]} & \multicolumn{1}{c}{\bf robust [95\% CI]} & \multicolumn{1}{c}{\bf intersection [95\% CI]} \\
        \hline \\
        L1 & haiku & s1-s2 & +0.000 [-0.113, +0.119] & -0.024 [-0.142, +0.097] & +0.006 [-0.109, +0.122] \\
        L1 & haiku & s1-s3 & +0.233 [+0.113, +0.352] & +0.225 [+0.108, +0.345] & +0.237 [+0.115, +0.359] \\
        L1 & haiku & s2-s3 & +0.233 [+0.138, +0.327] & +0.250 [+0.151, +0.351] & +0.231 [+0.135, +0.327] \\
        L1 & sonnet & s1-s2 & -0.069 [-0.176, +0.031] & -0.129 [-0.232, -0.030] & -0.071 [-0.173, +0.032] \\
        L1 & sonnet & s1-s3 & -0.069 [-0.176, +0.031] & -0.134 [-0.232, -0.041] & -0.077 [-0.179, +0.026] \\
        L1 & sonnet & s2-s3 & +0.000 [-0.069, +0.063] & -0.005 [-0.078, +0.064] & -0.006 [-0.077, +0.058] \\
        L1 & opus & s1-s2 & +0.069 [-0.013, +0.145] & +0.024 [-0.048, +0.095] & +0.064 [-0.013, +0.147] \\
        L1 & opus & s1-s3 & +0.126 [+0.031, +0.226] & -0.006 [-0.080, +0.069] & +0.115 [+0.019, +0.218] \\
        L1 & opus & s2-s3 & +0.057 [-0.006, +0.126] & -0.029 [-0.079, +0.017] & +0.051 [-0.013, +0.122] \\
        L1 & gemini-3.1-pro & s1-s2 & -0.044 [-0.138, +0.044] & -0.044 [-0.138, +0.044] & -0.045 [-0.141, +0.045] \\
        L1 & gemini-3.1-pro & s1-s3 & -0.057 [-0.151, +0.031] & -0.057 [-0.151, +0.031] & -0.064 [-0.154, +0.026] \\
        L1 & gemini-3.1-pro & s2-s3 & -0.013 [-0.107, +0.075] & -0.013 [-0.107, +0.075] & -0.019 [-0.109, +0.071] \\
        L1 & gpt-5 & s1-s2 & -0.031 [-0.113, +0.050] & -0.031 [-0.113, +0.050] & -0.051 [-0.128, +0.019] \\
        L1 & gpt-5 & s1-s3 & -0.038 [-0.094, +0.019] & -0.038 [-0.094, +0.019] & -0.051 [-0.103, +0.000] \\
        L1 & gpt-5 & s2-s3 & -0.006 [-0.063, +0.057] & -0.006 [-0.063, +0.057] & +0.000 [-0.051, +0.058] \\
        L2 & haiku & s1-s2 & -0.112 [-0.202, -0.023] & -0.193 [-0.286, -0.103] & -0.125 [-0.217, -0.033] \\
        L2 & haiku & s1-s3 & +0.089 [-0.004, +0.178] & +0.069 [-0.026, +0.161] & +0.096 [+0.000, +0.192] \\
        L2 & haiku & s2-s3 & +0.202 [+0.116, +0.287] & +0.262 [+0.168, +0.358] & +0.221 [+0.138, +0.304] \\
        L2 & sonnet & s1-s2 & -0.097 [-0.198, +0.004] & -0.221 [-0.331, -0.111] & -0.108 [-0.208, -0.004] \\
        L2 & sonnet & s1-s3 & -0.074 [-0.167, +0.019] & -0.186 [-0.284, -0.093] & -0.079 [-0.175, +0.012] \\
        L2 & sonnet & s2-s3 & +0.023 [-0.047, +0.097] & +0.035 [-0.043, +0.116] & +0.029 [-0.046, +0.104] \\
        L2 & opus & s1-s2 & -0.140 [-0.221, -0.062] & -0.279 [-0.352, -0.209] & -0.137 [-0.221, -0.054] \\
        L2 & opus & s1-s3 & -0.097 [-0.186, -0.012] & -0.266 [-0.348, -0.188] & -0.096 [-0.188, -0.004] \\
        L2 & opus & s2-s3 & +0.043 [-0.016, +0.105] & +0.013 [-0.035, +0.066] & +0.042 [-0.021, +0.108] \\
        L2 & gemini-3.1-pro & s1-s2 & -0.058 [-0.128, +0.012] & -0.058 [-0.128, +0.012] & -0.042 [-0.108, +0.025] \\
        L2 & gemini-3.1-pro & s1-s3 & -0.167 [-0.240, -0.093] & -0.167 [-0.240, -0.093] & -0.150 [-0.225, -0.079] \\
        L2 & gemini-3.1-pro & s2-s3 & -0.109 [-0.186, -0.035] & -0.109 [-0.186, -0.035] & -0.108 [-0.188, -0.033] \\
        L2 & gpt-5 & s1-s2 & -0.016 [-0.093, +0.058] & -0.016 [-0.093, +0.058] & -0.008 [-0.092, +0.075] \\
        L2 & gpt-5 & s1-s3 & -0.058 [-0.132, +0.016] & -0.058 [-0.132, +0.016] & -0.050 [-0.129, +0.029] \\
        L2 & gpt-5 & s2-s3 & -0.043 [-0.089, +0.008] & -0.043 [-0.089, +0.008] & -0.042 [-0.092, +0.008] \\
    \end{tabular}
    \end{center}
    \end{table}

\FloatBarrier

\section{Within-Anthropic tier sensitivity (RQ5)}\label{app:tier_sensitivity}
 
    This appendix reports the full RQ5 analysis: within the Claude 4.x family, does scaffold sensitivity scale predictably with model tier? The relevant quantity is the scaffold gap for each of Haiku, Sonnet, and Opus at each level under the three slices defined in Section~\ref{sec:scaffold_gap}, visualized in Figure~\ref{fig:scaffold_gap_bars} and tabulated in Appendix~\ref{app:scaffold_gap_full}. With only three tiers, monotonicity is read off directly rather than tested formally.
     
    Under the primary slice, the L1 gap ladder is non-monotonic: Haiku 0.23, Sonnet 0.07, Opus 0.13, with Opus exceeding Sonnet by 6 percentage points, the opposite direction from the H2 prediction. The L2 primary ladder follows the same shape (Haiku 0.20, Sonnet 0.10, Opus 0.14). On both levels, Sonnet is the least scaffold-sensitive Anthropic model, not Opus.
     
    The robust slice tells a different story at each level. On L1 the ladder is cleanly monotonic and decreasing (0.25, 0.13, 0.03): the H2 prediction holds once the SDK bug is removed, and Opus L1's 3-point gap is the smallest of any cell in the experiment. On L2 the robust ladder is flat and large (0.26, 0.22, 0.28): removing the bug expands all three Anthropic gaps and Opus emerges as the most scaffold-sensitive model. H2 fails completely at L2 under the robust slice; the most capable Anthropic model is the most scaffold-sensitive rather than the least.
     
    The intersection slice tracks the primary slice closely on both levels (52 of 53 L1 and 80 of 86 L2 sample\_ids retained) and does not change the qualitative reading; numbers are in Appendix~\ref{app:scaffold_gap_full}.
     
    Two interpretations of the L2 robust finding deserve attention. The first is that scaffold choice imposes larger absolute capability variation on more capable Anthropic models at the harder level (Opus L2's 28-point within-model spread is the experiment's largest): elicitation gets \emph{more} important at the top of the ladder when the task is hard enough that a poor scaffold leaves capability unused. The second is a ceiling-effect inversion: at L1 the better models have less room to move, compressing gaps at the top; at L2 they have more room, expanding them. The current data cannot adjudicate between these. What it can say is that the H2 prediction is conditionally supported on L1 under the robust slice and conditionally falsified on L2 under the same slice.

\section{Full GLMM coefficient tables and fit metadata}\label{app:glmm_full}

    This appendix reports the complete output for the pre-registered mixed-effects test of H2 (Section~\ref{sec:h2_inferential}): fit metadata for the baseline and interaction models under each slice, the likelihood-ratio test summary, and the full fixed-effect coefficient tables. All models are binomial GLMMs fit via lme4 (\texttt{glmer}) under the pymer4 wrapper with a random intercept on sample\_id. Reference levels are scaffold s1 (ReAct) and model Opus. The baseline formula is \texttt{success \textasciitilde{} scaffold + model + level + (1 | sample\_id)}; the interaction formula adds the scaffold $\times$ model interaction. The interaction fits report \texttt{converged = no} under lme4's strict gradient criterion; the gradient magnitudes at termination (0.001 to 0.013 across slices, discussed in Section~\ref{sec:h2_inferential}) are above the strict threshold but within practically converged range, and the fits are treated as usable for inference.

    \begin{table}[h]
        \caption{Fit metadata for the baseline and interaction GLMMs under each slice. n\_obs is attempt-units entering the fit; n\_ids is distinct sample\_ids contributing random intercepts.}
    \label{tab:glmm_fit_metadata}
    \begin{center}
    \small
    \begin{tabular}{llrrlrrr}
        \multicolumn{1}{c}{\bf slice} & \multicolumn{1}{c}{\bf model} & \multicolumn{1}{c}{\bf n\_obs} & \multicolumn{1}{c}{\bf n\_ids} & \multicolumn{1}{c}{\bf converged} & \multicolumn{1}{c}{\bf logLik} & \multicolumn{1}{c}{\bf AIC} & \multicolumn{1}{c}{\bf BIC} \\
        \hline \\
        primary & baseline & 6255 & 139 & yes & -2969.99 & 5957.98 & 6018.65 \\
        primary & interaction & 6255 & 139 & no & -2928.51 & 5891.03 & 6005.63 \\
        robust & baseline & 5941 & 139 & yes & -2611.75 & 5241.49 & 5301.70 \\
        robust & interaction & 5941 & 139 & no & -2550.66 & 5135.31 & 5249.04 \\
        intersection & baseline & 5940 & 132 & yes & -2807.72 & 5633.45 & 5693.65 \\
        intersection & interaction & 5940 & 132 & no & -2767.04 & 5568.07 & 5681.79 \\
    \end{tabular}
    \end{center}
    \end{table}

    \begin{table}[h]
        \caption{Likelihood-ratio test of the no-interaction null against the scaffold $\times$ model interaction model, per slice.}
    \label{tab:glmm_lrt}
    \begin{center}
    \small
    \begin{tabular}{lrrr}
        \multicolumn{1}{c}{\bf slice} & \multicolumn{1}{c}{\bf $\chi^2$} & \multicolumn{1}{c}{\bf df} & \multicolumn{1}{c}{\bf $p$} \\
        \hline \\
        primary & 82.95 & 8 & $<0.0001$ \\
        robust & 122.18 & 8 & $<0.0001$ \\
        intersection & 81.38 & 8 & $<0.0001$ \\
    \end{tabular}
    \end{center}
    \end{table}

    \begin{table}[h]
        \caption{Fixed-effect coefficients, baseline (no-interaction) GLMM, all three slices. Reference levels: scaffold s1, model Opus.}
    \label{tab:glmm_baseline_all}
    \begin{center}
    \footnotesize
    \begin{tabular}{lrrrr}
        \multicolumn{1}{c}{\bf term} & \multicolumn{1}{c}{\bf $\beta$} & \multicolumn{1}{c}{\bf SE} & \multicolumn{1}{c}{\bf 95\% CI} & \multicolumn{1}{c}{\bf $p$} \\
        \hline \\
        \multicolumn{5}{l}{\textit{Primary slice}} \\
        (Intercept)          &  1.701 & 0.445 & ( 0.829,  2.573) & $<0.001$ \\
        scaffold s2          &  0.404 & 0.082 & ( 0.244,  0.564) & $<0.001$ \\
        scaffold s3          &  0.156 & 0.080 & (-0.001,  0.314) & 0.052 \\
        model gemini-3.1-pro &  0.756 & 0.110 & ( 0.541,  0.971) & $<0.001$ \\
        model gpt-5          &  0.910 & 0.112 & ( 0.690,  1.129) & $<0.001$ \\
        model haiku          & -1.952 & 0.102 & (-2.152, -1.751) & $<0.001$ \\
        model sonnet         & -0.361 & 0.100 & (-0.557, -0.166) & $<0.001$ \\
        level                & -0.546 & 0.259 & (-1.052, -0.039) & 0.035 \\
        \hline \\
        \multicolumn{5}{l}{\textit{Robust slice}} \\
        (Intercept)          &  1.932 & 0.489 & ( 0.973,  2.890) & $<0.001$ \\
        scaffold s2          &  0.762 & 0.088 & ( 0.590,  0.935) & $<0.001$ \\
        scaffold s3          &  0.526 & 0.086 & ( 0.357,  0.695) & $<0.001$ \\
        model gemini-3.1-pro &  0.219 & 0.119 & (-0.015,  0.453) & 0.066 \\
        model gpt-5          &  0.379 & 0.121 & ( 0.142,  0.617) & 0.002 \\
        model haiku          & -2.520 & 0.118 & (-2.750, -2.289) & $<0.001$ \\
        model sonnet         & -0.587 & 0.115 & (-0.813, -0.362) & $<0.001$ \\
        level                & -0.449 & 0.284 & (-1.006,  0.108) & 0.114 \\
        \hline \\
        \multicolumn{5}{l}{\textit{Intersection slice}} \\
        (Intercept)          &  1.722 & 0.445 & ( 0.851,  2.594) & $<0.001$ \\
        scaffold s2          &  0.410 & 0.084 & ( 0.246,  0.575) & $<0.001$ \\
        scaffold s3          &  0.145 & 0.083 & (-0.017,  0.307) & 0.080 \\
        model gemini-3.1-pro &  0.634 & 0.113 & ( 0.413,  0.854) & $<0.001$ \\
        model gpt-5          &  0.855 & 0.116 & ( 0.628,  1.082) & $<0.001$ \\
        model haiku          & -2.002 & 0.106 & (-2.209, -1.795) & $<0.001$ \\
        model sonnet         & -0.366 & 0.103 & (-0.569, -0.163) & $<0.001$ \\
        level                & -0.497 & 0.260 & (-1.006,  0.013) & 0.056 \\
    \end{tabular}
    \end{center}
    \end{table}

    \begin{table}[h]
        \caption{Fixed-effect coefficients from the scaffold $\times$ model interaction GLMM, primary slice. Reference cell is Opus + s1. Interactions are log-odds relative to the s1+Opus baseline; positive means that scaffold helps that model more than it helps Opus. 95\% Wald CI shown. The robust and intersection slice coefficient tables follow in Table~\ref{tab:glmm_interaction_robust_intersection}.}
        \label{tab:h2_interaction_primary}
        \begin{center}
        \footnotesize
        \begin{tabular}{lrrrr}
            \multicolumn{1}{c}{\bf term} & \multicolumn{1}{c}{\bf $\beta$} & \multicolumn{1}{c}{\bf SE} & \multicolumn{1}{c}{\bf 95\% CI} & \multicolumn{1}{c}{\bf $p$} \\
            \hline \\
            (Intercept)                 &  1.752 & 0.461 & ( 0.849,  2.655) & $<0.001$ \\
            scaffold s2                 &  0.401 & 0.179 & ( 0.051,  0.751) & 0.025 \\
            scaffold s3                 &  0.077 & 0.174 & (-0.264,  0.418) & 0.659 \\
            model gemini-3.1-pro        &  0.452 & 0.179 & ( 0.101,  0.804) & 0.012 \\
            model gpt-5                 &  0.861 & 0.187 & ( 0.494,  1.227) & $<0.001$ \\
            model haiku                 & -1.663 & 0.171 & (-1.997, -1.328) & $<0.001$ \\
            model sonnet                & -0.519 & 0.169 & (-0.851, -0.187) & 0.002 \\
            level                       & -0.556 & 0.263 & (-1.071, -0.041) & 0.035 \\
            s2 $\times$ gemini-3.1-pro  &  0.008 & 0.262 & (-0.506,  0.522) & 0.975 \\
            s3 $\times$ gemini-3.1-pro  &  1.021 & 0.273 & ( 0.487,  1.555) & $<0.001$ \\
            s2 $\times$ gpt-5           & -0.214 & 0.270 & (-0.743,  0.315) & 0.427 \\
            s3 $\times$ gpt-5           &  0.384 & 0.272 & (-0.149,  0.917) & 0.158 \\
            s2 $\times$ haiku           & -0.015 & 0.242 & (-0.489,  0.459) & 0.949 \\
            s3 $\times$ haiku           & -0.972 & 0.247 & (-1.457, -0.487) & $<0.001$ \\
            s2 $\times$ sonnet          &  0.118 & 0.246 & (-0.364,  0.600) & 0.632 \\
            s3 $\times$ sonnet          &  0.351 & 0.242 & (-0.124,  0.825) & 0.148 \\
        \end{tabular}
        \end{center}
        \end{table}

    \begin{table}[h]
        \caption{Fixed-effect coefficients, scaffold $\times$ model interaction GLMM, robust and intersection slices. The primary-slice interaction table is Table~\ref{tab:h2_interaction_primary} above. Reference cell: Opus + s1.}
    \label{tab:glmm_interaction_robust_intersection}
    \begin{center}
    \footnotesize
    \begin{tabular}{lrrrr}
        \multicolumn{1}{c}{\bf term} & \multicolumn{1}{c}{\bf $\beta$} & \multicolumn{1}{c}{\bf SE} & \multicolumn{1}{c}{\bf 95\% CI} & \multicolumn{1}{c}{\bf $p$} \\
        \hline \\
        \multicolumn{5}{l}{\textit{Robust slice}} \\
        (Intercept)                &  1.654 & 0.509 & ( 0.657,  2.651) & 0.001 \\
        scaffold s2                &  1.197 & 0.207 & ( 0.790,  1.603) & $<0.001$ \\
        scaffold s3                &  1.285 & 0.215 & ( 0.864,  1.706) & $<0.001$ \\
        model gemini-3.1-pro       &  0.476 & 0.184 & ( 0.116,  0.836) & 0.010 \\
        model gpt-5                &  0.904 & 0.191 & ( 0.528,  1.279) & $<0.001$ \\
        model haiku                & -1.760 & 0.176 & (-2.105, -1.415) & $<0.001$ \\
        model sonnet               & -0.549 & 0.174 & (-0.890, -0.208) & 0.002 \\
        level                      & -0.466 & 0.292 & (-1.038,  0.106) & 0.110 \\
        s2 $\times$ gemini-3.1-pro & -0.768 & 0.285 & (-1.327, -0.210) & 0.007 \\
        s3 $\times$ gemini-3.1-pro & -0.138 & 0.302 & (-0.730,  0.453) & 0.647 \\
        s2 $\times$ gpt-5          & -1.001 & 0.293 & (-1.574, -0.428) & $<0.001$ \\
        s3 $\times$ gpt-5          & -0.803 & 0.302 & (-1.395, -0.211) & 0.008 \\
        s2 $\times$ haiku          & -0.531 & 0.269 & (-1.059, -0.003) & 0.049 \\
        s3 $\times$ haiku          & -2.129 & 0.282 & (-2.681, -1.577) & $<0.001$ \\
        s2 $\times$ sonnet         & -0.005 & 0.281 & (-0.556,  0.547) & 0.987 \\
        s3 $\times$ sonnet         & -0.230 & 0.285 & (-0.788,  0.328) & 0.419 \\
        \hline \\
        \multicolumn{5}{l}{\textit{Intersection slice}} \\
        (Intercept)                &  1.774 & 0.462 & ( 0.870,  2.679) & $<0.001$ \\
        scaffold s2                &  0.399 & 0.185 & ( 0.036,  0.762) & 0.031 \\
        scaffold s3                &  0.083 & 0.181 & (-0.271,  0.437) & 0.647 \\
        model gemini-3.1-pro       &  0.380 & 0.185 & ( 0.018,  0.743) & 0.040 \\
        model gpt-5                &  0.794 & 0.193 & ( 0.416,  1.173) & $<0.001$ \\
        model haiku                & -1.707 & 0.175 & (-2.051, -1.363) & $<0.001$ \\
        model sonnet               & -0.552 & 0.175 & (-0.894, -0.209) & 0.002 \\
        level                      & -0.508 & 0.264 & (-1.027,  0.010) & 0.055 \\
        s2 $\times$ gemini-3.1-pro & -0.068 & 0.270 & (-0.597,  0.460) & 0.800 \\
        s3 $\times$ gemini-3.1-pro &  0.931 & 0.280 & ( 0.383,  1.479) & $<0.001$ \\
        s2 $\times$ gpt-5          & -0.177 & 0.280 & (-0.726,  0.371) & 0.526 \\
        s3 $\times$ gpt-5          &  0.384 & 0.281 & (-0.168,  0.936) & 0.172 \\
        s2 $\times$ haiku          &  0.004 & 0.249 & (-0.484,  0.492) & 0.987 \\
        s3 $\times$ haiku          & -1.011 & 0.254 & (-1.509, -0.512) & $<0.001$ \\
        s2 $\times$ sonnet         &  0.171 & 0.255 & (-0.328,  0.670) & 0.502 \\
        s3 $\times$ sonnet         &  0.389 & 0.251 & (-0.102,  0.880) & 0.121 \\
    \end{tabular}
    \end{center}
    \end{table}

\section{H4 robustness checks}\label{app:h4_robustness}

    This appendix reports the full tables for the H4 difference-of-differences analysis (Section~\ref{sec:h4}). Table~\ref{tab:h4_dod} is the primary-slice DoD on the document-attachment subset; Table~\ref{tab:h4_robust} repeats it under the robust slice (\texttt{provider\_serialization\_bug} units excluded); Table~\ref{tab:h4_intersection} repeats it under the intersection slice; Table~\ref{tab:h4_anyfile} reports the primary-slice DoD under the expanded any-file definition, which adds image, audio, and code attachments to the document set (n = 11 L1, n = 20 L2 sample\_ids). Column definitions and the support/contradict reading are shared across all four tables.

    \begin{table}[h]
        \caption{H4 difference-of-differences (file subset minus non-file complement) for all three pairwise scaffold contrasts. Primary slice, document-attachment subset (n=6 L1, n=10 L2 sample\_ids). 95\% sample\_id-clustered bootstrap CI. A positive DoD on s3 minus s1 with CI strictly above zero supports H4; a negative DoD with CI strictly below zero contradicts it. One cell supports, four contradict, five inconclusive.}
        \label{tab:h4_dod}
        \begin{center}
        \small
        \begin{tabular}{llrrr}
            \multicolumn{1}{c}{\bf model} & \multicolumn{1}{c}{\bf level} & \multicolumn{1}{c}{\bf s3-s1 DoD [95\% CI]} & \multicolumn{1}{c}{\bf s3-s2 DoD [95\% CI]} & \multicolumn{1}{c}{\bf s2-s1 DoD [95\% CI]} \\
            \hline \\
            haiku          & 1 & -0.364 [-0.544, -0.172] & -0.364 [-0.636, -0.056] &  0.000 [-0.277,  0.262] \\
            sonnet         & 1 &  0.047 [-0.147,  0.324] &  0.063 [-0.052,  0.214] & -0.015 [-0.163,  0.156] \\
            opus           & 1 &  0.142 [ 0.030,  0.252] &  0.126 [ 0.015,  0.270] &  0.015 [-0.138,  0.136] \\
            gemini-3.1-pro & 1 & -0.001 [-0.136,  0.161] &  0.111 [-0.074,  0.388] & -0.112 [-0.275,  0.021] \\
            gpt-5          & 1 & -0.043 [-0.109,  0.021] & -0.007 [-0.068,  0.058] & -0.035 [-0.129,  0.061] \\
            haiku          & 2 & -0.314 [-0.531, -0.088] & -0.036 [-0.284,  0.228] & -0.278 [-0.424, -0.135] \\
            sonnet         & 2 &  0.030 [-0.230,  0.325] &  0.026 [-0.144,  0.223] &  0.004 [-0.343,  0.322] \\
            opus           & 2 &  0.079 [-0.073,  0.225] &  0.086 [-0.004,  0.190] & -0.007 [-0.143,  0.137] \\
            gemini-3.1-pro & 2 & -0.226 [-0.342, -0.133] & -0.161 [-0.278, -0.064] & -0.066 [-0.146,  0.013] \\
            gpt-5          & 2 & -0.141 [-0.310, -0.013] & -0.011 [-0.088,  0.083] & -0.131 [-0.375,  0.031] \\
        \end{tabular}
        \end{center}
    \end{table}

    \begin{table}[h]
        \caption{H4 difference-of-differences, document-attachment subset, robust slice. Zero of ten cells support H4 on the s3-s1 contrast; four contradict. The Opus L1 cell that supported H4 under the primary slice drops to -0.002.}
    \label{tab:h4_robust}
    \begin{center}
    \small
    \begin{tabular}{llrrr}
        \multicolumn{1}{c}{\bf model} & \multicolumn{1}{c}{\bf level} & \multicolumn{1}{c}{\bf s3-s1 DoD [95\% CI]} & \multicolumn{1}{c}{\bf s3-s2 DoD [95\% CI]} & \multicolumn{1}{c}{\bf s2-s1 DoD [95\% CI]} \\
        \hline \\
        haiku          & 1 & -0.361 [-0.541, -0.171] & -0.335 [-0.620, -0.046] & -0.026 [-0.296,  0.249] \\
        sonnet         & 1 & -0.023 [-0.214,  0.248] &  0.057 [-0.058,  0.204] & -0.081 [-0.223,  0.084] \\
        opus           & 1 & -0.002 [-0.083,  0.086] &  0.033 [-0.065,  0.175] & -0.035 [-0.186,  0.084] \\
        gemini-3.1-pro & 1 & -0.001 [-0.136,  0.161] &  0.111 [-0.074,  0.388] & -0.112 [-0.275,  0.021] \\
        gpt-5          & 1 & -0.043 [-0.109,  0.021] & -0.007 [-0.068,  0.058] & -0.035 [-0.129,  0.061] \\
        haiku          & 2 & -0.299 [-0.529, -0.051] &  0.002 [-0.287,  0.301] & -0.301 [-0.456, -0.145] \\
        sonnet         & 2 & -0.021 [-0.255,  0.241] &  0.111 [-0.027,  0.286] & -0.132 [-0.475,  0.190] \\
        opus           & 2 & -0.106 [-0.247,  0.034] &  0.055 [-0.021,  0.151] & -0.161 [-0.293, -0.031] \\
        gemini-3.1-pro & 2 & -0.226 [-0.342, -0.133] & -0.161 [-0.278, -0.064] & -0.066 [-0.146,  0.013] \\
        gpt-5          & 2 & -0.141 [-0.310, -0.013] & -0.011 [-0.087,  0.083] & -0.131 [-0.375,  0.031] \\
    \end{tabular}
    \end{center}
    \end{table}

    \begin{table}[h]
        \caption{H4 difference-of-differences, document-attachment subset, intersection slice. The Opus L1 supporting cell from the primary slice is restored; no additional supporting cells appear.}
    \label{tab:h4_intersection}
    \begin{center}
    \small
    \begin{tabular}{llrrr}
        \multicolumn{1}{c}{\bf model} & \multicolumn{1}{c}{\bf level} & \multicolumn{1}{c}{\bf s3-s1 DoD [95\% CI]} & \multicolumn{1}{c}{\bf s3-s2 DoD [95\% CI]} & \multicolumn{1}{c}{\bf s2-s1 DoD [95\% CI]} \\
        \hline \\
        haiku          & 1 & -0.360 [-0.543, -0.153] & -0.367 [-0.643, -0.054] &  0.007 [-0.264,  0.285] \\
        sonnet         & 1 &  0.039 [-0.157,  0.315] &  0.056 [-0.059,  0.200] & -0.017 [-0.167,  0.154] \\
        opus           & 1 &  0.130 [ 0.021,  0.241] &  0.121 [ 0.007,  0.271] &  0.010 [-0.145,  0.132] \\
        gemini-3.1-pro & 1 & -0.010 [-0.146,  0.154] &  0.104 [-0.083,  0.377] & -0.114 [-0.277,  0.025] \\
        gpt-5          & 1 & -0.058 [-0.116,  0.000] &  0.000 [-0.061,  0.067] & -0.058 [-0.146,  0.028] \\
        haiku          & 2 & -0.310 [-0.541, -0.067] & -0.014 [-0.271,  0.250] & -0.295 [-0.439, -0.149] \\
        sonnet         & 2 &  0.024 [-0.236,  0.310] &  0.033 [-0.138,  0.241] & -0.010 [-0.356,  0.304] \\
        opus           & 2 &  0.081 [-0.076,  0.234] &  0.086 [-0.005,  0.191] & -0.005 [-0.149,  0.144] \\
        gemini-3.1-pro & 2 & -0.210 [-0.324, -0.108] & -0.162 [-0.280, -0.061] & -0.048 [-0.124,  0.029] \\
        gpt-5          & 2 & -0.133 [-0.312, -0.005] & -0.010 [-0.089,  0.087] & -0.124 [-0.376,  0.041] \\
    \end{tabular}
    \end{center}
    \end{table}

    \begin{table}[h]
        \caption{H4 difference-of-differences, any-file subset (documents plus image, audio, and code attachments; n = 11 L1, n = 20 L2 sample\_ids), primary slice. One cell supports (Opus L1), three contradict (Haiku L1, Gemini L2, GPT-5 L2), six inconclusive.}
    \label{tab:h4_anyfile}
    \begin{center}
    \small
    \begin{tabular}{llrrr}
        \multicolumn{1}{c}{\bf model} & \multicolumn{1}{c}{\bf level} & \multicolumn{1}{c}{\bf s3-s1 DoD [95\% CI]} & \multicolumn{1}{c}{\bf s3-s2 DoD [95\% CI]} & \multicolumn{1}{c}{\bf s2-s1 DoD [95\% CI]} \\
        \hline \\
        haiku          & 1 & -0.242 [-0.454, -0.031] & -0.165 [-0.409,  0.079] & -0.076 [-0.322,  0.153] \\
        sonnet         & 1 &  0.027 [-0.178,  0.253] &  0.000 [-0.124,  0.117] &  0.027 [-0.136,  0.213] \\
        opus           & 1 &  0.273 [ 0.098,  0.497] &  0.110 [ 0.016,  0.216] &  0.164 [-0.012,  0.394] \\
        gemini-3.1-pro & 1 & -0.148 [-0.337,  0.022] & -0.092 [-0.335,  0.148] & -0.056 [-0.240,  0.142] \\
        gpt-5          & 1 & -0.048 [-0.120,  0.024] & -0.046 [-0.144,  0.041] & -0.001 [-0.116,  0.126] \\
        haiku          & 2 & -0.144 [-0.368,  0.094] &  0.045 [-0.161,  0.260] & -0.190 [-0.355, -0.013] \\
        sonnet         & 2 & -0.031 [-0.224,  0.164] &  0.009 [-0.171,  0.209] & -0.039 [-0.300,  0.216] \\
        opus           & 2 &  0.026 [-0.109,  0.159] &  0.121 [ 0.029,  0.218] & -0.095 [-0.234,  0.041] \\
        gemini-3.1-pro & 2 & -0.195 [-0.312, -0.077] & -0.120 [-0.273,  0.025] & -0.076 [-0.199,  0.056] \\
        gpt-5          & 2 & -0.184 [-0.335, -0.034] &  0.010 [-0.100,  0.123] & -0.194 [-0.358, -0.051] \\
    \end{tabular}
    \end{center}
    \end{table}

\section{Full behavioral signatures}\label{app:rq3_behavioral_full}

    This appendix reports the four behavioral signatures of RQ3 (action count, output tokens, working time on completed attempts, recovery-from-failure rate) across the three slices. One table per slice with both levels stacked; the L1 block is followed by a rule and the L2 block. CIs are 95\% sample\_id-clustered bootstrap, 5{,}000 resamples, seed shared with the rest of the validation analyses. Output tokens (thousands) and working time (seconds) are means. Recovery rate cells with no mid-trajectory errors in the denominator are reported as n/a; the (n recovered / n with error) raw counts are given in those cases. The full input/output/cache decomposition under each slice is in Appendix~\ref{app:rq3_tokens_split}.

    \begin{table}[h]
        \caption{Behavioral signatures, primary slice. CIs in parentheses; output tokens in thousands; working time in seconds (completed attempts only). L1 rows first, then L2 below the rule.}
    \label{tab:rq3_primary}
    \begin{center}
    \footnotesize
    \begin{tabular}{llrrrr}
        \multicolumn{1}{c}{\bf model} & \multicolumn{1}{c}{\bf scaf} & \multicolumn{1}{c}{\bf actions (95\% CI)} & \multicolumn{1}{c}{\bf out tok (k)} & \multicolumn{1}{c}{\bf time (s)} & \multicolumn{1}{c}{\bf recovery (n/N)} \\
        \hline \\
        gemini-3.1-pro & s1 & 15.8 (12.6-19.1) & 2.4 (1.8-3.1)  & 208 (168-251) & 0.00 (0/3) \\
        gpt-5          & s1 & 12.5 (9.2-16.1)  & 2.6 (1.8-3.6)  & 98 (69-132)   & 0.62 (30/48) \\
        haiku          & s1 & 22.6 (18.9-26.5) & 8.3 (6.2-10.7) & 124 (102-147) & 0.11 (4/36) \\
        opus           & s1 & 9.1 (6.1-12.3)   & 1.9 (1.1-3.0)  & 57 (38-79)    & 0.33 (1/3) \\
        sonnet         & s1 & 16.7 (12.7-20.9) & 5.1 (3.3-7.3)  & 136 (102-173) & 0.35 (7/20) \\
        gemini-3.1-pro & s2 & 10.1 (7.6-13.0)  & 1.7 (1.2-2.3)  & 159 (127-194) & n/a (0/0) \\
        gpt-5          & s2 & 4.7 (3.5-6.3)    & 1.8 (1.3-2.5)  & 76 (54-105)   & 0.80 (12/15) \\
        haiku          & s2 & 5.7 (3.7-8.3)    & 3.6 (2.2-5.3)  & 50 (35-68)    & n/a (0/0) \\
        opus           & s2 & 4.7 (2.8-7.1)    & 2.0 (1.0-3.2)  & 43 (21-72)    & 0.00 (0/6) \\
        sonnet         & s2 & 6.0 (4.1-8.3)    & 3.1 (1.7-5.0)  & 78 (50-114)   & 0.00 (0/5) \\
        gemini-3.1-pro & s3 & 6.1 (4.2-8.4)    & 1.4 (0.9-2.0)  & 124 (94-160)  & n/a (0/0) \\
        gpt-5          & s3 & 2.3 (1.0-4.1)    & 1.8 (1.2-2.7)  & 73 (50-103)   & 0.62 (8/13) \\
        haiku          & s3 & 4.2 (2.2-6.8)    & 3.1 (1.8-4.7)  & 37 (24-54)    & n/a (0/0) \\
        opus           & s3 & 4.4 (2.4-7.0)    & 2.4 (1.5-3.7)  & 62 (39-91)    & 0.00 (0/5) \\
        sonnet         & s3 & 4.8 (2.9-7.1)    & 2.6 (1.4-4.1)  & 62 (39-92)    & 0.00 (0/1) \\
        \hline \\
        gemini-3.1-pro & s1 & 26.6 (23.7-29.6) & 3.7 (3.2-4.1)  & 255 (221-288) & 0.50 (4/8) \\
        gpt-5          & s1 & 16.3 (13.8-19.0) & 3.5 (2.9-4.0)  & 121 (98-145)  & 0.65 (70/108) \\
        haiku          & s1 & 29.9 (26.9-32.7) & 8.7 (7.4-10.2) & 152 (135-170) & 0.19 (13/68) \\
        opus           & s1 & 12.9 (10.3-15.5) & 2.4 (1.8-3.1)  & 55 (33-79)    & 0.29 (6/21) \\
        sonnet         & s1 & 24.2 (20.6-27.9) & 5.8 (4.6-7.2)  & 176 (146-206) & 0.28 (12/43) \\
        gemini-3.1-pro & s2 & 13.6 (11.2-16.0) & 2.4 (1.9-2.9)  & 209 (174-245) & n/a (0/0) \\
        gpt-5          & s2 & 8.8 (7.3-10.4)   & 3.4 (2.7-4.2)  & 119 (97-143)  & 0.66 (35/53) \\
        haiku          & s2 & 6.4 (4.8-8.4)    & 3.8 (2.8-5.1)  & 62 (48-80)    & 1.00 (1/1) \\
        opus           & s2 & 7.6 (5.7-9.7)    & 2.4 (1.7-3.3)  & 77 (57-101)   & 0.29 (5/17) \\
        sonnet         & s2 & 7.6 (5.7-9.6)    & 3.5 (2.5-4.7)  & 106 (78-137)  & 0.33 (6/18) \\
        gemini-3.1-pro & s3 & 10.0 (7.8-12.5)  & 2.0 (1.6-2.5)  & 153 (125-183) & n/a (0/0) \\
        gpt-5          & s3 & 4.8 (3.6-6.0)    & 2.9 (2.3-3.6)  & 87 (72-103)   & 0.75 (30/40) \\
        haiku          & s3 & 4.4 (3.0-6.2)    & 3.0 (2.2-4.0)  & 44 (33-58)    & 0.33 (1/3) \\
        opus           & s3 & 6.9 (5.0-9.4)    & 2.9 (2.1-4.1)  & 83 (63-108)   & 0.59 (10/17) \\
        sonnet         & s3 & 6.1 (4.5-7.9)    & 3.6 (2.6-4.8)  & 98 (72-130)   & 0.38 (5/13) \\
    \end{tabular}
    \end{center}
    \end{table}

\FloatBarrier

    \begin{table}[h]
        \caption{Behavioral signatures, robust slice (no flag). L1 rows first, then L2 below the rule. Same conventions as Table~\ref{tab:rq3_primary}.}
    \label{tab:rq3_robust}
    \begin{center}
    \footnotesize
    \begin{tabular}{llrrrr}
        \multicolumn{1}{c}{\bf model} & \multicolumn{1}{c}{\bf scaf} & \multicolumn{1}{c}{\bf actions (95\% CI)} & \multicolumn{1}{c}{\bf out tok (k)} & \multicolumn{1}{c}{\bf time (s)} & \multicolumn{1}{c}{\bf recovery (n/N)} \\
        \hline \\
        gemini-3.1-pro & s1 & 15.8 (12.6-19.1) & 2.4 (1.8-3.1)  & 208 (168-251) & 0.00 (0/3) \\
        gpt-5          & s1 & 12.5 (9.2-16.1)  & 2.6 (1.8-3.6)  & 98 (69-132)   & 0.62 (30/48) \\
        haiku          & s1 & 22.6 (18.9-26.5) & 8.3 (6.2-10.6) & 124 (102-147) & 0.11 (4/36) \\
        opus           & s1 & 9.1 (6.2-12.2)   & 1.9 (1.1-3.0)  & 57 (38-80)    & 0.33 (1/3) \\
        sonnet         & s1 & 16.7 (12.7-21.0) & 5.1 (3.3-7.3)  & 136 (102-173) & 0.35 (7/20) \\
        gemini-3.1-pro & s2 & 10.1 (7.6-13.0)  & 1.7 (1.2-2.3)  & 159 (127-194) & n/a (0/0) \\
        gpt-5          & s2 & 4.7 (3.5-6.3)    & 1.8 (1.3-2.5)  & 76 (54-105)   & 0.80 (12/15) \\
        haiku          & s2 & 5.9 (3.9-8.6)    & 3.7 (2.2-5.5)  & 50 (34-68)    & n/a (0/0) \\
        opus           & s2 & 4.6 (2.9-6.9)    & 1.9 (1.0-3.2)  & 43 (20-73)    & 0.00 (0/5) \\
        sonnet         & s2 & 6.4 (4.4-8.8)    & 3.2 (1.7-5.3)  & 78 (50-114)   & 0.00 (0/5) \\
        gemini-3.1-pro & s3 & 6.1 (4.2-8.4)    & 1.4 (0.9-2.0)  & 124 (94-160)  & n/a (0/0) \\
        gpt-5          & s3 & 2.3 (1.0-4.1)    & 1.8 (1.2-2.7)  & 73 (50-103)   & 0.62 (8/13) \\
        haiku          & s3 & 4.3 (2.2-6.9)    & 3.2 (1.9-4.8)  & 37 (24-54)    & n/a (0/0) \\
        opus           & s3 & 4.8 (2.6-7.6)    & 2.5 (1.5-3.8)  & 62 (39-91)    & 0.00 (0/5) \\
        sonnet         & s3 & 5.1 (3.1-7.6)    & 2.7 (1.5-4.4)  & 62 (39-92)    & 0.00 (0/1) \\
        \hline \\
        gemini-3.1-pro & s1 & 26.6 (23.7-29.6) & 3.7 (3.2-4.1)  & 255 (221-288) & 0.50 (4/8) \\
        gpt-5          & s1 & 16.3 (13.8-19.0) & 3.5 (2.9-4.0)  & 121 (98-145)  & 0.65 (70/108) \\
        haiku          & s1 & 29.9 (26.9-32.7) & 8.7 (7.4-10.2) & 152 (135-170) & 0.19 (13/68) \\
        opus           & s1 & 12.9 (10.3-15.5) & 2.4 (1.8-3.1)  & 55 (33-79)    & 0.29 (6/21) \\
        sonnet         & s1 & 24.2 (20.6-27.9) & 5.8 (4.6-7.2)  & 176 (146-206) & 0.28 (12/43) \\
        gemini-3.1-pro & s2 & 13.6 (11.2-16.0) & 2.4 (1.9-2.9)  & 209 (174-245) & n/a (0/0) \\
        gpt-5          & s2 & 8.8 (7.3-10.5)   & 3.4 (2.7-4.2)  & 119 (98-143)  & 0.66 (35/53) \\
        haiku          & s2 & 7.4 (5.4-9.7)    & 4.2 (3.0-5.7)  & 62 (48-80)    & 1.00 (1/1) \\
        opus           & s2 & 8.5 (6.3-11.0)   & 2.6 (1.8-3.7)  & 77 (57-101)   & 0.36 (5/14) \\
        sonnet         & s2 & 8.6 (6.5-11.0)   & 3.9 (2.6-5.4)  & 106 (78-137)  & 0.40 (6/15) \\
        gemini-3.1-pro & s3 & 10.0 (7.8-12.5)  & 2.0 (1.6-2.5)  & 153 (125-183) & n/a (0/0) \\
        gpt-5          & s3 & 4.8 (3.6-6.0)    & 2.9 (2.3-3.6)  & 87 (72-103)   & 0.75 (30/40) \\
        haiku          & s3 & 4.7 (3.1-6.6)    & 3.1 (2.2-4.1)  & 44 (33-57)    & 0.50 (1/2) \\
        opus           & s3 & 8.3 (5.9-11.1)   & 3.3 (2.3-4.7)  & 83 (63-107)   & 0.59 (10/17) \\
        sonnet         & s3 & 6.9 (5.0-9.1)    & 3.9 (2.7-5.3)  & 98 (72-130)   & 0.42 (5/12) \\
    \end{tabular}
    \end{center}
    \end{table}

\FloatBarrier

    \begin{table}[h]
        \caption{Behavioral signatures, intersection slice. L1 rows first, then L2 below the rule. Same conventions as Table~\ref{tab:rq3_primary}.}
    \label{tab:rq3_intersection}
    \begin{center}
    \footnotesize
    \begin{tabular}{llrrrr}
        \multicolumn{1}{c}{\bf model} & \multicolumn{1}{c}{\bf scaf} & \multicolumn{1}{c}{\bf actions (95\% CI)} & \multicolumn{1}{c}{\bf out tok (k)} & \multicolumn{1}{c}{\bf time (s)} & \multicolumn{1}{c}{\bf recovery (n/N)} \\
        \hline \\
        gemini-3.1-pro & s1 & 15.9 (12.6-19.4) & 2.5 (1.8-3.2)  & 210 (169-253) & 0.00 (0/3) \\
        gpt-5          & s1 & 12.2 (9.0-16.0)  & 2.6 (1.8-3.5)  & 97 (67-132)   & 0.61 (28/46) \\
        haiku          & s1 & 22.5 (18.7-26.3) & 8.3 (6.1-10.6) & 124 (101-146) & 0.12 (4/34) \\
        opus           & s1 & 9.0 (6.2-12.3)   & 1.9 (1.0-3.0)  & 56 (37-80)    & 0.33 (1/3) \\
        sonnet         & s1 & 16.5 (12.4-20.8) & 5.1 (3.2-7.4)  & 135 (100-174) & 0.39 (7/18) \\
        gemini-3.1-pro & s2 & 10.0 (7.5-13.0)  & 1.7 (1.2-2.3)  & 158 (125-194) & n/a (0/0) \\
        gpt-5          & s2 & 4.8 (3.5-6.3)    & 1.7 (1.2-2.4)  & 76 (54-104)   & 0.80 (12/15) \\
        haiku          & s2 & 5.8 (3.7-8.2)    & 3.6 (2.2-5.3)  & 50 (34-69)    & n/a (0/0) \\
        opus           & s2 & 4.7 (2.9-7.1)    & 1.9 (1.0-3.2)  & 43 (20-71)    & 0.00 (0/6) \\
        sonnet         & s2 & 5.9 (4.0-8.3)    & 2.9 (1.4-4.8)  & 71 (45-104)   & 0.00 (0/5) \\
        gemini-3.1-pro & s3 & 5.9 (4.0-8.3)    & 1.3 (0.9-1.9)  & 121 (91-156)  & n/a (0/0) \\
        gpt-5          & s3 & 2.4 (0.9-4.2)    & 1.8 (1.1-2.6)  & 73 (48-102)   & 0.62 (8/13) \\
        haiku          & s3 & 4.2 (2.2-6.8)    & 3.1 (1.8-4.7)  & 37 (24-54)    & n/a (0/0) \\
        opus           & s3 & 4.4 (2.4-7.0)    & 2.4 (1.4-3.7)  & 61 (39-90)    & 0.00 (0/5) \\
        sonnet         & s3 & 4.9 (3.0-7.1)    & 2.6 (1.5-4.1)  & 62 (39-90)    & 0.00 (0/1) \\
        \hline \\
        gemini-3.1-pro & s1 & 26.0 (23.0-29.0) & 3.6 (3.1-4.1)  & 252 (218-286) & 0.57 (4/7) \\
        gpt-5          & s1 & 16.0 (13.4-18.7) & 3.5 (2.9-4.1)  & 121 (98-147)  & 0.67 (65/97) \\
        haiku          & s1 & 29.5 (26.4-32.5) & 8.7 (7.3-10.2) & 149 (132-168) & 0.20 (13/65) \\
        opus           & s1 & 12.7 (10.2-15.6) & 2.5 (1.9-3.2)  & 57 (34-83)    & 0.30 (6/20) \\
        sonnet         & s1 & 23.8 (20.2-27.4) & 5.9 (4.6-7.4)  & 175 (144-209) & 0.31 (12/39) \\
        gemini-3.1-pro & s2 & 14.0 (11.5-16.7) & 2.5 (2.0-3.0)  & 212 (176-250) & n/a (0/0) \\
        gpt-5          & s2 & 8.6 (7.0-10.5)   & 3.4 (2.7-4.2)  & 118 (95-142)  & 0.66 (31/47) \\
        haiku          & s2 & 6.9 (5.1-9.0)    & 4.0 (2.9-5.3)  & 63 (48-80)    & 1.00 (1/1) \\
        opus           & s2 & 7.9 (5.9-10.2)   & 2.5 (1.7-3.5)  & 79 (56-106)   & 0.29 (5/17) \\
        sonnet         & s2 & 8.0 (6.0-10.2)   & 3.6 (2.5-4.9)  & 106 (76-138)  & 0.33 (6/18) \\
        gemini-3.1-pro & s3 & 10.5 (8.1-13.0)  & 2.1 (1.6-2.7)  & 157 (127-189) & n/a (0/0) \\
        gpt-5          & s3 & 4.8 (3.6-6.1)    & 2.9 (2.3-3.7)  & 86 (70-103)   & 0.76 (29/38) \\
        haiku          & s3 & 4.6 (3.0-6.5)    & 3.1 (2.2-4.1)  & 44 (33-57)    & 0.33 (1/3) \\
        opus           & s3 & 7.1 (5.1-9.5)    & 3.0 (2.1-4.2)  & 83 (62-109)   & 0.59 (10/17) \\
        sonnet         & s3 & 6.4 (4.7-8.3)    & 3.6 (2.6-4.9)  & 97 (70-128)   & 0.38 (5/13) \\
    \end{tabular}
    \end{center}
    \end{table}

\FloatBarrier

\section{Token decomposition: input, output, cache-read, cache-write}\label{app:rq3_tokens_split}

    The body of the paper uses output tokens as the cross-provider behavioral signature, since provider-reported \texttt{total\_tokens} mixes the input, output, cache-read, and cache-write streams and the cache accounting varies by provider. This appendix reports all four streams per cell (means in thousands of tokens), allowing the reader to read the full token activity behind each (model, scaffold, level) configuration. Three notes: (a) Anthropic models report cache-read separately from input and frequently show input=0.0 on s1 because the prompt is identical across attempts and the second-and-later attempts hit the cache for the full prompt. (b) OpenAI models do not surface a separate cache-write field, so cache-write reads as 0.0 for GPT-5. (c) Gemini's total includes a small unaccounted-for residual (typically 4--7k tokens per attempt), which is consistent with hosted-tool token activity (the \texttt{google\_search} grounding tool) that does not surface in the four named streams. All tables are from the primary slice; the robust and intersection slices are available in the supplementary materials.

    \begin{table}[h]
        \caption{Token decomposition (thousands of tokens, primary slice). Rows are (model, scaffold), L1 stacked above L2 below the rule. The Total column is the provider-reported \texttt{total\_tokens}; for Gemini, Total exceeds the sum of the four streams by a small residual (hosted-tool tokens).}
    \label{tab:rq3_tokens_primary}
    \begin{center}
    \footnotesize
    \begin{tabular}{llrrrrr}
        \multicolumn{1}{c}{\bf model} & \multicolumn{1}{c}{\bf scaf} & \multicolumn{1}{c}{\bf input} & \multicolumn{1}{c}{\bf output} & \multicolumn{1}{c}{\bf cache rd} & \multicolumn{1}{c}{\bf cache wr} & \multicolumn{1}{c}{\bf total} \\
        \hline \\
        gemini-3.1-pro & s1 & 61.9  & 2.4 & 179.3  & 0.0  & 248.9  \\
        gpt-5          & s1 & 23.5  & 2.6 & 160.4  & 0.0  & 186.6  \\
        haiku          & s1 & 7.8   & 8.3 & 777.5  & 44.0 & 837.6  \\
        opus           & s1 & 0.0   & 1.9 & 143.4  & 12.8 & 158.1  \\
        sonnet         & s1 & 0.0   & 5.1 & 394.9  & 25.5 & 425.6  \\
        gemini-3.1-pro & s2 & 60.3  & 1.7 & 57.7   & 0.0  & 126.4  \\
        gpt-5          & s2 & 18.6  & 1.8 & 38.5   & 0.0  & 58.9   \\
        haiku          & s2 & 7.4   & 3.6 & 185.0  & 66.9 & 263.0  \\
        opus           & s2 & 8.0   & 2.0 & 102.9  & 44.5 & 157.3  \\
        sonnet         & s2 & 12.8  & 3.1 & 255.2  & 92.2 & 363.2  \\
        gemini-3.1-pro & s3 & 34.6  & 1.4 & 34.6   & 0.0  & 74.5   \\
        gpt-5          & s3 & 14.7  & 1.8 & 41.5   & 0.0  & 58.0   \\
        haiku          & s3 & 5.7   & 3.1 & 103.6  & 30.2 & 142.6  \\
        opus           & s3 & 4.0   & 2.4 & 97.9   & 22.9 & 127.2  \\
        sonnet         & s3 & 4.5   & 2.6 & 141.9  & 27.7 & 176.8  \\
        \hline \\
        gemini-3.1-pro & s1 & 99.6  & 3.7 & 367.2  & 0.0   & 476.2  \\
        gpt-5          & s1 & 34.3  & 3.5 & 222.9  & 0.0   & 260.7  \\
        haiku          & s1 & 7.5   & 8.7 & 1256.4 & 65.9  & 1338.5 \\
        opus           & s1 & 0.0   & 2.4 & 294.0  & 17.3  & 313.8  \\
        sonnet         & s1 & 0.0   & 5.8 & 675.7  & 36.8  & 718.4  \\
        gemini-3.1-pro & s2 & 83.2  & 2.4 & 88.5   & 0.0   & 180.8  \\
        gpt-5          & s2 & 41.8  & 3.4 & 105.4  & 0.0   & 150.6  \\
        haiku          & s2 & 8.6   & 3.8 & 456.6  & 132.4 & 601.3  \\
        opus           & s2 & 10.0  & 2.4 & 242.8  & 59.0  & 314.2  \\
        sonnet         & s2 & 23.1  & 3.5 & 481.0  & 118.7 & 626.3  \\
        gemini-3.1-pro & s3 & 55.4  & 2.0 & 64.5   & 0.0   & 126.4  \\
        gpt-5          & s3 & 24.1  & 2.9 & 58.3   & 0.0   & 85.3   \\
        haiku          & s3 & 7.1   & 3.0 & 226.0  & 58.2  & 294.2  \\
        opus           & s3 & 12.8  & 2.9 & 231.8  & 35.6  & 283.1  \\
        sonnet         & s3 & 25.5  & 3.6 & 530.3  & 84.4  & 643.7  \\
    \end{tabular}
    \end{center}
    \end{table}

\section{Full cost-per-correct ranking}\label{app:cost_per_correct_full}

    This appendix reports the per-level realized cost-per-correct rankings under the three slices (primary, robust no-flag, intersection). Each row is one (model, scaffold) cell at the indicated level. Rank is ascending by realized cost per correct, so rank 1 is the cheapest cell on the panel. The point estimate and 95\% bootstrap CI come from 5{,}000 sample\_id-clustered resamples (seed shared with Sections~\ref{sec:bootstrap_accuracy}, \ref{sec:scaffold_gap_stat}, and~\ref{sec:cost_per_correct}). The slice definitions match the rest of the paper: primary includes every attempt-unit; robust excludes units carrying the \texttt{provider\_serialization\_bug} flag; intersection retains only sample\_ids that completed cleanly in every (model, scaffold) cell at the level.

    \subsection{Primary slice}\label{app:cpc_primary}

        \begin{table}[t]
            \caption{Realized cost per correct, primary slice, Level 1. Rank is ascending. n\_attempted is 159 for every cell at L1. cpc realized is total log-derived realized spend on the cell divided by n\_correct.}
        \label{tab:cpc_primary_L1_ranked}
        \begin{center}
        \small
        \begin{tabular}{rllrrr}
            \multicolumn{1}{c}{\bf rank} & \multicolumn{1}{c}{\bf model} & \multicolumn{1}{c}{\bf scaffold} & \multicolumn{1}{c}{\bf n\_correct} & \multicolumn{1}{c}{\bf total cost (USD)} & \multicolumn{1}{c}{\bf cpc realized [95\% CI]} \\
            \hline \\
            1 & gemini-3.1-pro & s3 & 134 & 18.57 & 0.139 [0.062, 0.257] \\
            2 & gpt-5 & s3 & 135 & 23.56 & 0.175 [0.107, 0.276] \\
            3 & gpt-5 & s2 & 134 & 26.26 & 0.196 [0.139, 0.274] \\
            4 & opus & s1 & 132 & 31.72 & 0.240 [0.153, 0.354] \\
            5 & gemini-3.1-pro & s1 & 125 & 31.08 & 0.249 [0.174, 0.350] \\
            6 & gemini-3.1-pro & s2 & 132 & 33.36 & 0.253 [0.134, 0.429] \\
            7 & sonnet & s3 & 116 & 31.67 & 0.273 [0.177, 0.398] \\
            8 & haiku & s2 & 73 & 22.03 & 0.302 [0.199, 0.463] \\
            9 & haiku & s3 & 36 & 11.02 & 0.306 [0.206, 0.484] \\
            10 & gpt-5 & s1 & 129 & 44.17 & 0.342 [0.222, 0.497] \\
            11 & opus & s3 & 112 & 43.36 & 0.387 [0.255, 0.574] \\
            12 & haiku & s1 & 73 & 28.93 & 0.396 [0.277, 0.568] \\
            13 & sonnet & s1 & 105 & 46.31 & 0.441 [0.296, 0.646] \\
            14 & opus & s2 & 121 & 66.61 & 0.550 [0.346, 0.807] \\
            15 & sonnet & s2 & 116 & 80.58 & 0.695 [0.317, 1.334] \\
        \end{tabular}
        \end{center}
        \end{table}
        
        \begin{table}[t]
            \caption{Realized cost per correct, primary slice, Level 2. Rank is ascending. n\_attempted is 258 for every cell at L2.}
        \label{tab:cpc_primary_L2_ranked}
        \begin{center}
        \small
        \begin{tabular}{rllrrr}
            \multicolumn{1}{c}{\bf rank} & \multicolumn{1}{c}{\bf model} & \multicolumn{1}{c}{\bf scaffold} & \multicolumn{1}{c}{\bf n\_correct} & \multicolumn{1}{c}{\bf total cost (USD)} & \multicolumn{1}{c}{\bf cpc realized [95\% CI]} \\
            \hline \\
            1 & gemini-3.1-pro & s3 & 222 & 50.79 & 0.229 [0.145, 0.330] \\
            2 & gpt-5 & s3 & 212 & 61.09 & 0.288 [0.221, 0.378] \\
            3 & gemini-3.1-pro & s2 & 194 & 79.03 & 0.407 [0.279, 0.563] \\
            4 & gemini-3.1-pro & s1 & 179 & 87.98 & 0.491 [0.381, 0.639] \\
            5 & gpt-5 & s1 & 197 & 99.86 & 0.507 [0.386, 0.662] \\
            6 & gpt-5 & s2 & 201 & 101.98 & 0.507 [0.362, 0.694] \\
            7 & haiku & s3 & 58 & 30.26 & 0.522 [0.382, 0.755] \\
            8 & opus & s1 & 144 & 81.57 & 0.566 [0.403, 0.791] \\
            9 & haiku & s2 & 110 & 71.52 & 0.650 [0.451, 0.925] \\
            10 & opus & s3 & 169 & 122.71 & 0.726 [0.519, 0.983] \\
            11 & sonnet & s1 & 135 & 110.60 & 0.819 [0.610, 1.133] \\
            12 & haiku & s1 & 81 & 66.96 & 0.827 [0.616, 1.157] \\
            13 & opus & s2 & 180 & 155.06 & 0.861 [0.633, 1.156] \\
            14 & sonnet & s3 & 154 & 156.19 & 1.014 [0.684, 1.448] \\
            15 & sonnet & s2 & 160 & 191.80 & 1.199 [0.818, 1.695] \\
        \end{tabular}
        \end{center}
        \end{table}

    \FloatBarrier
    \subsection{Robust slice (\texttt{provider\_serialization\_bug} units excluded)}\label{app:cpc_robust}
    
        \begin{table}[t]
            \caption{Realized cost per correct, robust slice, Level 1. Rank is ascending. Cells unaffected by the flag (cross-provider models, Anthropic L1 cells without flagged units) match the primary L1 values; flagged units removed for Anthropic cells where they appear.}
        \label{tab:cpc_robust_L1_ranked}
        \begin{center}
        \small
        \begin{tabular}{rllrrr}
            \multicolumn{1}{c}{\bf rank} & \multicolumn{1}{c}{\bf model} & \multicolumn{1}{c}{\bf scaffold} & \multicolumn{1}{c}{\bf n\_correct} & \multicolumn{1}{c}{\bf total cost (USD)} & \multicolumn{1}{c}{\bf cpc realized [95\% CI]} \\
            \hline \\
            1 & gemini-3.1-pro & s3 & 134 & 18.57 & 0.139 [0.062, 0.257] \\
            2 & gpt-5 & s3 & 135 & 23.56 & 0.175 [0.107, 0.276] \\
            3 & gpt-5 & s2 & 134 & 26.26 & 0.196 [0.139, 0.274] \\
            4 & opus & s1 & 132 & 31.72 & 0.240 [0.153, 0.354] \\
            5 & gemini-3.1-pro & s1 & 125 & 31.08 & 0.249 [0.174, 0.350] \\
            6 & gemini-3.1-pro & s2 & 132 & 33.36 & 0.253 [0.134, 0.429] \\
            7 & sonnet & s3 & 116 & 31.67 & 0.273 [0.177, 0.398] \\
            8 & haiku & s2 & 73 & 21.07 & 0.289 [0.193, 0.441] \\
            9 & haiku & s3 & 36 & 11.02 & 0.306 [0.206, 0.484] \\
            10 & gpt-5 & s1 & 129 & 44.17 & 0.342 [0.222, 0.497] \\
            11 & opus & s3 & 112 & 38.13 & 0.340 [0.226, 0.508] \\
            12 & haiku & s1 & 73 & 28.93 & 0.396 [0.277, 0.568] \\
            13 & sonnet & s1 & 105 & 46.31 & 0.441 [0.296, 0.646] \\
            14 & opus & s2 & 121 & 60.61 & 0.501 [0.319, 0.741] \\
            15 & sonnet & s2 & 116 & 77.50 & 0.668 [0.305, 1.282] \\
        \end{tabular}
        \end{center}
        \end{table}
        
        \begin{table}[t]
            \caption{Realized cost per correct, robust slice, Level 2. Rank is ascending. The Anthropic L2 cells carrying flagged units (Opus, Sonnet, and Haiku s2/s3) drop relative to the primary L2 ranking; cross-provider cells are unchanged. Ranks 6 and 7 coincide at three decimals; ordering is by full-precision cpc.}
        \label{tab:cpc_robust_L2_ranked}
        \begin{center}
        \small
        \begin{tabular}{rllrrr}
            \multicolumn{1}{c}{\bf rank} & \multicolumn{1}{c}{\bf model} & \multicolumn{1}{c}{\bf scaffold} & \multicolumn{1}{c}{\bf n\_correct} & \multicolumn{1}{c}{\bf total cost (USD)} & \multicolumn{1}{c}{\bf cpc realized [95\% CI]} \\
            \hline \\
            1 & gemini-3.1-pro & s3 & 222 & 50.79 & 0.229 [0.145, 0.330] \\
            2 & gpt-5 & s3 & 212 & 61.09 & 0.288 [0.221, 0.378] \\
            3 & gemini-3.1-pro & s2 & 194 & 79.03 & 0.407 [0.279, 0.563] \\
            4 & haiku & s3 & 58 & 27.30 & 0.471 [0.347, 0.669] \\
            5 & gemini-3.1-pro & s1 & 179 & 87.98 & 0.491 [0.381, 0.639] \\
            6 & gpt-5 & s1 & 197 & 99.86 & 0.507 [0.386, 0.662] \\
            7 & gpt-5 & s2 & 201 & 101.98 & 0.507 [0.364, 0.706] \\
            8 & haiku & s2 & 110 & 60.95 & 0.554 [0.400, 0.775] \\
            9 & opus & s1 & 144 & 81.57 & 0.566 [0.403, 0.791] \\
            10 & opus & s3 & 169 & 100.49 & 0.595 [0.414, 0.811] \\
            11 & opus & s2 & 180 & 135.50 & 0.753 [0.549, 1.005] \\
            12 & sonnet & s1 & 135 & 110.60 & 0.819 [0.610, 1.133] \\
            13 & haiku & s1 & 81 & 66.96 & 0.827 [0.616, 1.157] \\
            14 & sonnet & s3 & 154 & 127.94 & 0.831 [0.574, 1.158] \\
            15 & sonnet & s2 & 160 & 155.16 & 0.970 [0.668, 1.348] \\
        \end{tabular}
        \end{center}
        \end{table}

    \FloatBarrier
    \subsection{Intersection slice (per-level shared sample\_ids)}\label{app:cpc_intersection}

    \begin{table}[t]
            \caption{Realized cost per correct, intersection slice, Level 1. Rank is ascending. The intersection retains 52 of 53 sample\_ids at L1; n\_attempted is 156 for every cell.}
        \label{tab:cpc_intersection_L1_ranked}
        \begin{center}
        \small
        \begin{tabular}{rllrrr}
            \multicolumn{1}{c}{\bf rank} & \multicolumn{1}{c}{\bf model} & \multicolumn{1}{c}{\bf scaffold} & \multicolumn{1}{c}{\bf n\_correct} & \multicolumn{1}{c}{\bf total cost (USD)} & \multicolumn{1}{c}{\bf cpc realized [95\% CI]} \\
            \hline \\
            1 & gemini-3.1-pro & s3 & 132 & 17.92 & 0.136 [0.059, 0.259] \\
            2 & gpt-5 & s3 & 134 & 22.76 & 0.170 [0.103, 0.274] \\
            3 & gpt-5 & s2 & 134 & 25.63 & 0.191 [0.133, 0.270] \\
            4 & opus & s1 & 129 & 30.20 & 0.234 [0.146, 0.347] \\
            5 & gemini-3.1-pro & s1 & 122 & 30.71 & 0.252 [0.174, 0.354] \\
            6 & gemini-3.1-pro & s2 & 129 & 32.65 & 0.253 [0.133, 0.430] \\
            7 & sonnet & s3 & 116 & 31.33 & 0.270 [0.175, 0.397] \\
            8 & haiku & s3 & 36 & 10.28 & 0.286 [0.194, 0.448] \\
            9 & haiku & s2 & 72 & 20.74 & 0.288 [0.188, 0.443] \\
            10 & gpt-5 & s1 & 126 & 42.59 & 0.338 [0.217, 0.496] \\
            11 & opus & s3 & 111 & 40.81 & 0.368 [0.239, 0.541] \\
            12 & haiku & s1 & 73 & 28.31 & 0.388 [0.267, 0.567] \\
            13 & sonnet & s1 & 104 & 44.58 & 0.429 [0.286, 0.626] \\
            14 & sonnet & s2 & 115 & 53.13 & 0.462 [0.271, 0.721] \\
            15 & opus & s2 & 119 & 63.17 & 0.531 [0.330, 0.778] \\
        \end{tabular}
        \end{center}
        \end{table}
        
        \begin{table}[t]
            \caption{Realized cost per correct, intersection slice, Level 2. Rank is ascending. The intersection retains 80 of 86 sample\_ids at L2; n\_attempted is 240 for every cell.}
        \label{tab:cpc_intersection_L2_ranked}
        \begin{center}
        \small
        \begin{tabular}{rllrrr}
            \multicolumn{1}{c}{\bf rank} & \multicolumn{1}{c}{\bf model} & \multicolumn{1}{c}{\bf scaffold} & \multicolumn{1}{c}{\bf n\_correct} & \multicolumn{1}{c}{\bf total cost (USD)} & \multicolumn{1}{c}{\bf cpc realized [95\% CI]} \\
            \hline \\
            1 & gemini-3.1-pro & s3 & 205 & 49.87 & 0.243 [0.154, 0.354] \\
            2 & gpt-5 & s3 & 198 & 56.37 & 0.285 [0.214, 0.376] \\
            3 & gemini-3.1-pro & s2 & 179 & 77.36 & 0.432 [0.294, 0.605] \\
            4 & gemini-3.1-pro & s1 & 169 & 80.46 & 0.476 [0.366, 0.619] \\
            5 & gpt-5 & s1 & 186 & 92.17 & 0.496 [0.376, 0.652] \\
            6 & gpt-5 & s2 & 188 & 93.94 & 0.500 [0.351, 0.710] \\
            7 & haiku & s3 & 54 & 28.30 & 0.524 [0.377, 0.767] \\
            8 & opus & s1 & 140 & 75.03 & 0.536 [0.380, 0.746] \\
            9 & haiku & s2 & 107 & 69.07 & 0.645 [0.448, 0.933] \\
            10 & opus & s3 & 163 & 115.16 & 0.706 [0.495, 0.967] \\
            11 & sonnet & s1 & 129 & 102.28 & 0.793 [0.579, 1.094] \\
            12 & haiku & s1 & 77 & 61.43 & 0.798 [0.586, 1.138] \\
            13 & opus & s2 & 173 & 148.32 & 0.857 [0.625, 1.157] \\
            14 & sonnet & s3 & 148 & 127.14 & 0.859 [0.591, 1.206] \\
            15 & sonnet & s2 & 155 & 147.37 & 0.951 [0.676, 1.311] \\
        \end{tabular}
        \end{center}
        \end{table}

\FloatBarrier

\section{Realized versus happy-path cost per correct}\label{app:cost_per_correct_happy_path}

    This appendix reports realized and happy-path cost per correct side-by-side for the primary slice. Happy-path cost per correct uses each completed unit's latest-eval cost (rather than the sum across all retries) and restricts both numerator and denominator to completed units only. The two metrics coincide for any cell with 100\% completion, since there are no retries and no errored units to differentiate. They diverge where errored units carried token cost without contributing to the n\_correct denominator; the gap quantifies the cost overhead of attempts that paid for tokens but did not score.

    \begin{table}[t]
        \caption{Realized versus happy-path cost per correct, primary slice, Level 1. Sorted ascending by realized. Cells with realized == happy-path are 100\% completion cells; cells with realized > happy-path carry token cost on errored units.}
    \label{tab:cpc_happy_L1}
    \begin{center}
    \small
    \begin{tabular}{llrrr}
        \multicolumn{1}{c}{\bf model} & \multicolumn{1}{c}{\bf scaffold} & \multicolumn{1}{c}{\bf cpc realized} & \multicolumn{1}{c}{\bf cpc happy-path} & \multicolumn{1}{c}{\bf delta} \\
        \hline \\
        gemini-3.1-pro & s3 & 0.139 & 0.139 & 0.000 \\
        gpt-5 & s3 & 0.175 & 0.175 & 0.000 \\
        gpt-5 & s2 & 0.196 & 0.196 & 0.000 \\
        opus & s1 & 0.240 & 0.240 & 0.000 \\
        gemini-3.1-pro & s1 & 0.249 & 0.249 & 0.000 \\
        gemini-3.1-pro & s2 & 0.253 & 0.253 & 0.000 \\
        sonnet & s3 & 0.273 & 0.273 & 0.000 \\
        haiku & s2 & 0.302 & 0.289 & 0.013 \\
        haiku & s3 & 0.306 & 0.306 & 0.000 \\
        gpt-5 & s1 & 0.342 & 0.342 & 0.000 \\
        opus & s3 & 0.387 & 0.340 & 0.047 \\
        haiku & s1 & 0.396 & 0.396 & 0.000 \\
        sonnet & s1 & 0.441 & 0.441 & 0.000 \\
        opus & s2 & 0.550 & 0.501 & 0.049 \\
        sonnet & s2 & 0.695 & 0.668 & 0.027 \\
    \end{tabular}
    \end{center}
    \end{table}
    
    \begin{table}[t]
        \caption{Realized versus happy-path cost per correct, primary slice, Level 2. Sorted ascending by realized. The Anthropic L2 cluster shows the largest deltas, driven by \texttt{provider\_serialization\_bug} units that paid token cost without completing.}
    \label{tab:cpc_happy_L2}
    \begin{center}
    \small
    \begin{tabular}{llrrr}
        \multicolumn{1}{c}{\bf model} & \multicolumn{1}{c}{\bf scaffold} & \multicolumn{1}{c}{\bf cpc realized} & \multicolumn{1}{c}{\bf cpc happy-path} & \multicolumn{1}{c}{\bf delta} \\
        \hline \\
        gemini-3.1-pro & s3 & 0.229 & 0.229 & 0.000 \\
        gpt-5 & s3 & 0.288 & 0.288 & 0.000 \\
        gemini-3.1-pro & s2 & 0.407 & 0.407 & 0.000 \\
        gemini-3.1-pro & s1 & 0.491 & 0.491 & 0.000 \\
        gpt-5 & s1 & 0.507 & 0.507 & 0.000 \\
        gpt-5 & s2 & 0.507 & 0.507 & 0.000 \\
        haiku & s3 & 0.522 & 0.470 & 0.052 \\
        opus & s1 & 0.566 & 0.566 & 0.000 \\
        haiku & s2 & 0.650 & 0.487 & 0.163 \\
        opus & s3 & 0.726 & 0.595 & 0.131 \\
        sonnet & s1 & 0.819 & 0.819 & 0.000 \\
        haiku & s1 & 0.827 & 0.824 & 0.003 \\
        opus & s2 & 0.861 & 0.753 & 0.108 \\
        sonnet & s3 & 1.014 & 0.831 & 0.183 \\
        sonnet & s2 & 1.199 & 0.970 & 0.229 \\
    \end{tabular}
    \end{center}
    \end{table}

\FloatBarrier

\section{\texttt{provider\_serialization\_bug} per-cell summary}
\label{app:provider_bug_summary}

    Sample-level counts of records carrying the \texttt{provider\_serialization\_bug} flag, by (model, scaffold, attempt) cell. The flag fires when Anthropic's API returned a \texttt{web\_fetch\_tool\_result} error code that the published Anthropic Python SDK could not deserialize. No flagged sample received a clean retry; the resume sweep targeted credit-exhaustion failures, not flagged samples.
    
    \begin{table}[t]
    \caption{Per-cell counts of \texttt{provider\_serialization\_bug}-flagged samples in the validation run. 314 samples across 13 cells. concentration is on scaffolds s2 and s3, consistent with web\_fetch usage patterns in those scaffolds.}
    \label{tab:provider_bug_summary}
    \begin{center}
    \begin{tabular}{llrrrr}
        \multicolumn{1}{c}{\bf model} &\multicolumn{1}{c}{\bf scaffold} &\multicolumn{1}{c}{\bf attempt} &\multicolumn{1}{c}{\bf n\_flagged} &\multicolumn{1}{c}{\bf n\_resolved} &\multicolumn{1}{c}{\bf n\_unresolved} \\
        \hline \\
        haiku &s2 &1 &4 &0 &4 \\
        haiku &s2 &2 &9 &0 &9 \\
        haiku &s2 &3 &36 &0 &36 \\
        haiku &s3 &3 &26 &0 &26 \\
        opus &s2 &2 &18 &0 &18 \\
        opus &s2 &3 &34 &0 &34 \\
        opus &s3 &2 &37 &0 &37 \\
        opus &s3 &3 &41 &0 &41 \\
        sonnet &s2 &2 &2 &0 &2 \\
        sonnet &s2 &3 &53 &0 &53 \\
        sonnet &s3 &1 &1 &0 &1 \\
        sonnet &s3 &2 &1 &0 &1 \\
        sonnet &s3 &3 &52 &0 &52 \\
    \end{tabular}
    \end{center}
    \end{table}

\section{\texttt{prompt\_too\_long} retry-threading table}
\label{app:prompt_too_long_threading} 

    Each row corresponds to a unique (sample\_id, model, scaffold, attempt) unit that hit a \texttt{prompt is too long} error in at least one \texttt{.eval} file. \texttt{later\_flagged} is True if a subsequent \texttt{.eval} for the same unit was assigned the \texttt{provider\_serialization\_bug} flag. \texttt{later\_resolved} is True if a subsequent \texttt{.eval} completed without error.

    \begin{table}[t]
        \caption{\texttt{prompt is too long} retry-threading for the validation run. 33 units (39 raw error records). 15 units later flagged with \texttt{provider\_serialization\_bug}; 3 later resolved cleanly; 15 with neither (13 remained terminal, the \texttt{p2l} bucket in Table~\ref{tab:per_cell_outcomes_full}; 2 ended in a different error). sample IDs truncated to first 8 characters; full IDs are in the project repository.}
    \label{tab:prompt_too_long_threading}
    \begin{center}
    \begin{tabular}{llrrcc}
        \multicolumn{1}{c}{\bf sample\_id} &\multicolumn{1}{c}{\bf model} &\multicolumn{1}{c}{\bf scaffold} &\multicolumn{1}{c}{\bf attempt} &\multicolumn{1}{c}{\bf later\_flagged} &\multicolumn{1}{c}{\bf later\_resolved} \\
        \hline \\
        3ff6b7a9 &haiku &s2 &1 &Y & \\
        48eb8242 &haiku &s2 &1 &Y & \\
        65638e28 &haiku &s2 &1 &Y & \\
        71345b0a &haiku &s2 &1 & &Y \\
        7673d772 &haiku &s2 &1 &Y & \\
        7a4a336d &haiku &s2 &1 & & \\
        9f41b083 &haiku &s2 &1 & & \\
        c8b7e059 &haiku &s2 &1 & & \\
        08c0b6e9 &haiku &s2 &2 & &Y \\
        0a3cd321 &haiku &s2 &2 &Y & \\
        0b260a57 &haiku &s2 &2 & &Y \\
        16d825ff &haiku &s2 &2 &Y & \\
        1dcc160f &haiku &s2 &2 &Y & \\
        3ff6b7a9 &haiku &s2 &2 &Y & \\
        48eb8242 &haiku &s2 &2 &Y & \\
        4b6bb5f7 &haiku &s2 &2 &Y & \\
        65638e28 &haiku &s2 &2 &Y & \\
        7a4a336d &haiku &s2 &2 &Y & \\
        9f41b083 &haiku &s2 &2 & & \\
        c8b7e059 &haiku &s2 &2 & & \\
        ecbc4f94 &haiku &s2 &2 &Y & \\
        56137764 &haiku &s2 &3 & & \\
        7673d772 &haiku &s2 &3 & & \\
        9f41b083 &haiku &s2 &3 & & \\
        9f41b083 &haiku &s3 &1 & & \\
        9f41b083 &haiku &s3 &2 & & \\
        65638e28 &haiku &s3 &3 & & \\
        9f41b083 &haiku &s3 &3 & & \\
        48eb8242 &opus &s2 &3 & & \\
        48eb8242 &sonnet &s2 &2 &Y & \\
        c8b7e059 &sonnet &s2 &2 &Y & \\
        48eb8242 &sonnet &s3 &2 & & \\
        48eb8242 &sonnet &s3 &3 & & \\
    \end{tabular}
    \end{center}
    \end{table}

\FloatBarrier

\section{Content-filter block inspection}\label{app:content_filter_inspection}

    28 sample-level records carried the \texttt{Output blocked by content filtering policy} error across two unique GAIA questions. The 28 raw records span 21 distinct attempt-units: 15 terminally blocked (the \texttt{cf} bucket in Table~\ref{tab:per_cell_outcomes_full}), 4 resolved on a later clean run, and 2 that ended in a different error. Each block below shows the GAIA question text followed by the (model, scaffold, attempt, level) cells in which the block occurred, with turn counts at the time of the block.

    \subsection{Question A:}
    
    \begin{table}[H]
        \caption{content-filter block: question A. 23 occurrences across the Anthropic models. All blocked on the same level 2 GAIA question. Several rows duplicate because the sample was retried within the same attempt and was blocked again.}
    \label{tab:content_filter_question_a}
    \begin{center}
    \begin{tabular}{|p{.14\linewidth} p{.14\linewidth} p{.14\linewidth} p{.14\linewidth} p{.14\linewidth} p{.14\linewidth}|}
        \hline
        \multicolumn{6}{|l|}{\textbf{sample\_id:}\rule[-1.2ex]{0pt}{0pt}} \\
        \multicolumn{6}{|l|}{ed58682d-bc52-4baa-9eb0-4eb81e1edacc\rule[-1.2ex]{0pt}{0pt}} \\
        \hline
        \multicolumn{6}{|l|}{\textbf{Question:}\rule[-1.2ex]{0pt}{0pt}} \\
        \multicolumn{6}{|p{\linewidth-2\tabcolsep-2\arrayrulewidth}|}{What is the last word before the second chorus of the King of Pop's fifth single from his sixth studio album?\rule[-1.2ex]{0pt}{0pt}} \\
        \hline
        \rule[-1.2ex]{0pt}{0pt}
        \textbf{model} & \textbf{scaffold} & \textbf{attempt} & \textbf{n\_total} & \textbf{n\_assistant} & \textbf{n\_tool} \\
        \hline 
        haiku &s1 &1 &16 &7 &7 \\
        haiku &s1 &2 &16 &7 &7 \\
        haiku &s1 &2 &24 &11 &11 \\
        haiku &s2 &2 &3 &1 &0 \\
        haiku &s3 &1 &3 &1 &0 \\
        haiku &s3 &1 &3 &1 &0 \\
        haiku &s3 &2 &3 &1 &0 \\
        haiku &s3 &2 &3 &1 &0 \\
        haiku &s3 &3 &3 &1 &0 \\
        opus &s1 &3 &6 &2 &2 \\
        opus &s2 &2 &3 &1 &0 \\
        opus &s2 &3 &3 &1 &0 \\
        sonnet &s1 &1 &8 &3 &3 \\
        sonnet &s2 &1 &1 &0 &0 \\
        sonnet &s2 &1 &1 &0 &0 \\
        sonnet &s2 &2 &1 &0 &0 \\
        sonnet &s2 &2 &1 &0 &0 \\
        sonnet &s2 &3 &1 &0 &0 \\
        sonnet &s3 &1 &3 &1 &0 \\
        sonnet &s3 &1 &3 &1 &0 \\
        sonnet &s3 &2 &1 &0 &0 \\
        sonnet &s3 &2 &1 &0 &0 \\
        sonnet &s3 &3 &3 &1 &0 \\
        \hline
    \end{tabular}
    \end{center}
    \end{table}
    
   \newpage
    \subsection{Question B:}

    \begin{table}[!h]
        \caption{content-filter block: question B. 5 occurrences, all sonnet on the same level 1 GAIA question. Block fired both on initial user message (3-turn rows) and mid-trajectory after tool calls (12-turn and 56-turn rows).}
    \label{tab:content_filter_question_b}
    \begin{center}
    \begin{tabular}{|p{.14\linewidth} p{.14\linewidth} p{.14\linewidth} p{.14\linewidth} p{.14\linewidth} p{.14\linewidth}|}
        \hline
        \multicolumn{6}{|l|}{\textbf{sample\_id:}\rule[-1.2ex]{0pt}{0pt}} \\
        \multicolumn{6}{|l|}{7673d772-ef80-4f0f-a602-1bf4485c9b43\rule[-1.2ex]{0pt}{0pt}} \\
        \hline
        \multicolumn{6}{|l|}{\textbf{Question:}\rule[-1.2ex]{0pt}{0pt}} \\
        \multicolumn{6}{|p{\linewidth-2\tabcolsep-2\arrayrulewidth}|}{On Cornell Law School website's legal information institute, under the fifth section of federal rules alphabetically, what word was deleted in the last amendment to the first rule in the article that has "witnesses" in the most titles as of 2021?\rule[-1.2ex]{0pt}{0pt}} \\
        \hline
        \textbf{model} & \textbf{scaffold} & \textbf{attempt} & \textbf{n\_total} & \textbf{n\_assistant} & \textbf{n\_tool} \\
        \hline 
        sonnet &s1 &2 &12 &5 &5 \\
        sonnet &s1 &3 &56 &27 &27 \\
        sonnet &s3 &1 &3 &1 &0 \\
        sonnet &s3 &2 &5 &2 &1 \\
        sonnet &s3 &3 &3 &1 &0 \\
        \hline
    \end{tabular}
    \end{center}
    \end{table}

\end{document}